\pdfoutput=1
\documentclass[letterpaper, 10 pt, journal, twoside]{./support/IEEEtran}

\usepackage{times}
\usepackage[pdftex]{graphicx}
\usepackage{subfigure}
\usepackage{amsmath,amssymb,amsopn,amstext,amsfonts}
\usepackage{cancel}
\usepackage[space]{cite}
\usepackage{soul}
\usepackage{cuted}
\usepackage{lipsum,multicol}
\usepackage{stfloats}
\everymath{\displaystyle}
\usepackage{booktabs}
\usepackage{threeparttable}
\usepackage{amsthm}
\usepackage{xfrac}
\usepackage{balance}
\usepackage{color}
\usepackage{mathtools}

\usepackage{tabularx}
\usepackage{makecell}
\usepackage{bm}
\usepackage{diagbox}
\usepackage{float}
\usepackage{epstopdf}
\usepackage{url}
\usepackage{multirow}
\usepackage{multicol}
\usepackage{tikz}
\usepackage{hyperref}[linkcolor=black,citecolor=black,urlcolor=black,colorlinks=true]
\usepackage{enumitem}
\usepackage[ruled,vlined,linesnumbered]{algorithm2e}
\usepackage{pifont}
\usepackage{siunitx}
\usepackage{diagbox}
\usepackage{array}
\usepackage{xcolor}
\DeclareMathAlphabet\mathbfcal{OMS}{cmsy}{b}{n}

\let\oldFootnote\footnote
\newcommand\nextToken\relax

\renewcommand\footnote[1]{%
    \oldFootnote{#1}\futurelet\nextToken\isFootnote}

\newcommand\isFootnote{%
    \ifx\footnote\nextToken\textsuperscript{,}\fi}

\newcommand{\etal}{\textit{et al}. }
\newcommand{\ie}{\textit{i}.\textit{e}.}
\newcommand{\eg}{\textit{e}.\textit{g}.}

\definecolor{celadon}{rgb}{0.78, 0.93, 0.80}
\pagecolor{celadon}
\nopagecolor
\soulregister\cite7
\soulregister\citep7
\soulregister\citet7
\soulregister\ref7
\soulregister\it7
\soulregister\pageref7

\usepackage[numbers]{natbib}
\usepackage[resetlabels]{multibib}
\newcites{sec}{Reference}

\usepackage{arydshln}
\graphicspath{{figures/}}
\DeclareGraphicsExtensions{.pdf,.png,.jpg,.eps}

\IEEEoverridecommandlockouts

\title{Swarm-LIO2: Decentralized, Efficient LiDAR-inertial Odometry for UAV Swarms}

\author{Fangcheng Zhu$^{*}$, Yunfan Ren$^{*}$, Longji Yin$^{*}$, Fanze Kong, Qingbo Liu,\\ Ruize Xue, Wenyi Liu, Yixi Cai, Guozheng Lu, Haotian Li, Fu Zhang$^\dag$
    \thanks{$^{*}$Equal contribution.}
    \thanks{$^\dag$Corresponding author: Fu Zhang.}
	\thanks{All authors are with the Mechatronics and Robotic Systems (MaRS) Laboratory, Department of Mechanical Engineering, The University of Hong Kong, Hong Kong SAR, China. 
 (email: \{zhufc, renyf, ljyin, kongfz, liuwenyi, yixicai, gzlu, haotianl\}@connect.hku.hk;  \{qingboliu0703,xrz836884211\}@gmail.com; fuzhang@hku.hk)
 }
}

\begin{document}
\maketitle
\begin{abstract}
Aerial swarm systems possess immense potential in various aspects, such as cooperative exploration, target tracking, search and rescue. Efficient, accurate self and mutual state estimation are the critical preconditions for completing these swarm tasks, which remain challenging research topics. This paper proposes Swarm-LIO2: a fully decentralized, plug-and-play, computationally efficient, and bandwidth-efficient LiDAR-inertial odometry for aerial swarm systems. 
Swarm-LIO2 uses a decentralized, plug-and-play network as the communication infrastructure. Only bandwidth-efficient and low-dimensional information is exchanged, including identity, ego-state, mutual observation measurements, and global extrinsic transformations. To support the plug-and-play of new teammate participants, Swarm-LIO2 detects potential teammate UAVs and initializes the temporal offset and global extrinsic transformation all automatically. To enhance the initialization efficiency, novel reflectivity-based UAV detection, trajectory matching, and factor graph optimization methods are proposed. For state estimation, Swarm-LIO2 fuses LiDAR, IMU, and mutual observation measurements within an efficient ESIKF framework, with careful compensation of temporal delay and modeling of measurements to enhance the accuracy and consistency. Moreover, the proposed ESIKF framework leverages the global extrinsic for ego-state estimation in case of LiDAR degeneration or refines the global extrinsic along with the ego-state estimation otherwise. To enhance the scalability, Swarm-LIO2 introduces a novel marginalization method in the ESIKF, which \textcolor{black}{prevents the growth of computational time with} swarm size. 
Extensive simulation and real-world experiments demonstrate the broad adaptability to large-scale aerial swarm systems and complicated scenarios, including GPS-denied scenes, degenerated scenes for cameras or LiDARs.
The experimental results showcase the centimeter-level localization accuracy which outperforms other state-of-the-art LiDAR-inertial odometry for a single UAV system.
Furthermore, diverse applications demonstrate the potential of Swarm-LIO2 to serve as reliable infrastructure for various aerial swarm missions. In addition, we open-source all the system designs on GitHub to benefit society: \href{https://github.com/hku-mars/Swarm-LIO2}{\tt github.com/hku-mars/Swarm-LIO2}.
\end{abstract}

\section{Introduction}\label{section:intro}
\IEEEPARstart{I}{n} recent years, multi-robot system\textcolor{black}{s}, especially aerial swarm system\textcolor{black}{s}, ha\textcolor{black}{ve exhibited} great potential in many fields, such as collaborative autonomous exploration\cite{gao2022meeting, zhou2023racer, tang2023bubble}, target tracking\cite{zhu2020multi, bonatti2020autonomous,olfati2011collaborative,swarm_tracking}, search and rescue\cite{arnold2018search,queralta2020collaborative,li2023collaborative}, etc. Thanks to \textcolor{black}{their} great team cooperation capability, swarm systems can complete various missions in complex scenarios, even in degenerated environments for a single robot.
For a single robot system, well-developed state estimation techniques provide accurate ego-state estimation \cite{xu2021fast,xu2022fast,qin2018vins,lin2022r,zhang2014loam}, serving as a critical precondition for a wide variety of autonomous tasks such as trajectory planning\cite{ren2022bubble,kong2021avoiding,ren2022online} and motion control\cite{lu2022model}. For robotic swarm systems, state estimation plays \textcolor{black}{an equally} significant role\cite{xu2020decentralized,xu2022omni}, where each robot \textcolor{black}{needs to estimate the state of the self UAV (\ie, ego-state estimation) as well as the other teammate UAVs (\ie, mutual state estimation)}.  \textcolor{black}{Accurate and robust estimation of ego and mutual states is crucial for the robot swarms to collaborate on a task. }

\begin{figure}[t]
	\setlength\abovecaptionskip{-0.1\baselineskip}
	\centering
	\includegraphics[width=\linewidth]{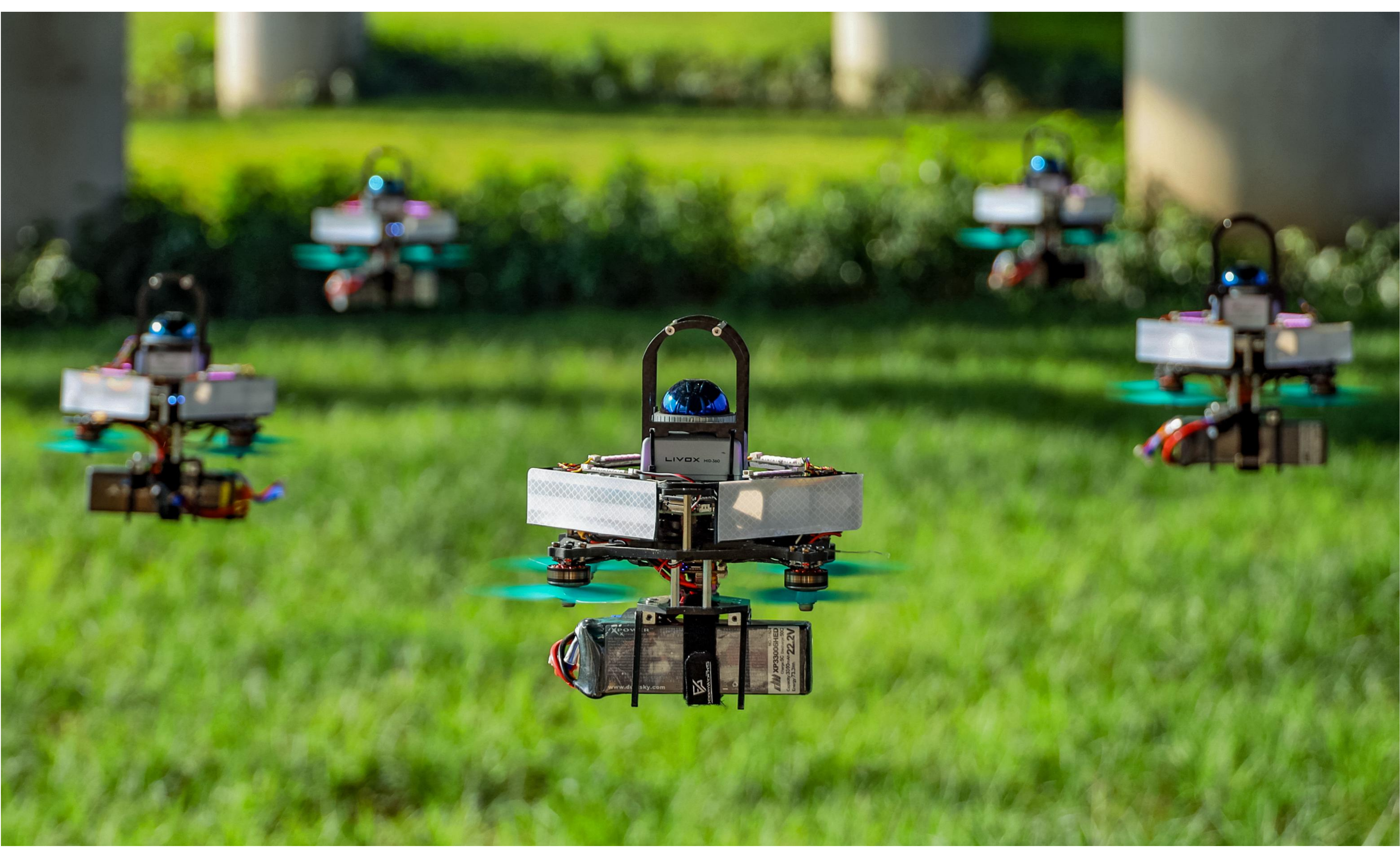}
	\caption{An aerial swarm system constituted of five UAVs flying in the wild in which Swarm-LIO2 serves as the self and mutual state estimator. More details can be found in the attached video at \href{ https://youtu.be/Q7cJ9iRhlrY}{\textcolor{black}{https://youtu.be/Q7cJ9iRhlrY}}}
	\label{fig:cover} 
  \vspace{-5mm}
\end{figure} 

\textcolor{black}{Over the past few decades}, multiple sensors and devices \textcolor{black}{have been} adopted to achieve reliable state estimation for robotic swarm systems.
GPS and RTK-GPS are \textcolor{black}{commonly used for self-localization in outdoor environments, as reported in previous studies \cite{jaimes2008approach, moon2016outdoor}}. For GPS-denied environments, motion capture systems \cite{honig2018trajectory} and anchor-based Ultra-WideBand (UWB) systems \cite{ledergerber2015robot,zhou2021online,zhou2022swarm} \textcolor{black}{have been utilized for state estimation in multi-robot systems}. \textcolor{black}{These methods \cite{honig2018trajectory,ledergerber2015robot,zhou2021online,zhou2022swarm} often rely on the stationary ground station, resulting in a centralized system that is prone to single-point-of-failure. In more recent research, cameras have become a popular choice in multi-robot systems due to their lightweight design, low cost, and rich color information. These camera-based systems are often complemented by an Inertial Measurement Unit (IMU) and an anchor-free UWB to provide more robust state estimation \cite{zhou2021ego, schmuck2021covins, weinstein2018visual, zhou2022swarm, xu2022omni, guo2019ultra}.}
However, cameras are vulnerable to inadequate illumination and lack direct depth measurements, leading to high computational complexity in computing 3D measurements.
Although the complementary anchor-free UWB can provide distance measurements, \textcolor{black}{it is susceptible to multi-path effects and obstacle occlusion in the environment}, which decreases the overall system accuracy. 

\textcolor{black}{In recent years}, 3-D light detection and ranging (LiDAR) sensors \textcolor{black}{have gained popularity in state estimation due to their ability to provide direct, accurate 3D measurements over a long range and various illumination conditions.} While traditional mechanical spinning LiDARs are often expensive and heavy, \textcolor{black}{recent advancements in LiDAR technology have introduced cost-effective and lightweight LiDARs that are suitable for deployment on mobile robots, particularly unmanned aerial vehicles (UAVs).} These LiDARs have not only enabled the development of autonomous navigation systems using LiDARs for UAVs \cite{ren2023rog,cai2023occupancy,chen2023self,chen2023swashplateless}, but opened up new possibilities for state estimation in swarm systems.

\textcolor{black}{Leveraging the above LiDAR advantages, this paper aims to develop a fully decentralized, plug-and-play, computationally efficient, and bandwidth-friendly state estimation method for aerial swarm systems based on LiDAR-inertial measurements. Fully decentralized means no master agent exists in any module of the whole system from communication hardware to algorithm software, which avoids single-point-of-failure. Plug-and-play means that an agent can automatically join the swarm and easily collaborate with other teammates before or in the middle of a mission. The system must also be computationally efficient and bandwidth-efficient, since the limited payload capacity of UAVs imposes significant constraints on the computational resources and network bandwidth.}

\textcolor{black}{We decompose the task of aerial swarm state estimation into two key online modules, initialization and state estimation.
In the initialization module, each UAV needs to detect and identify possible new teammate UAVs, and calibrate temporal offsets and global extrinsic transformations with the found teammates.
To achieve simple but effective teammate detection, reflective tapes are attached to each UAV, making teammate UAVs easily detectable from LiDAR reflectivity measurements. This teammate detection is conducted in real-time at each LiDAR scan measurement, enabling the detection of new teammate UAVs even in the middle of a mission. For each detected new UAV, its identification and global extrinsic transformation are obtained through trajectory matching, while the global extrinsic with the rest of teammate UAVs found on the network are swiftly calibrated through a factor graph optimization. Moreover, by exchanging low-dimensional data via a decentralized Ad-Hoc network, teammate monitoring and temporal calibration can be fulfilled efficiently and in a fully decentralized manner. }

In the state estimation module, each UAV in the swarm system performs real-time, robust, and precise ego-state estimation \textcolor{black}{as well as mutual state estimation. Estimating the full state of all teammates in each UAV is computationally demanding. Thus, we propose to estimate on each UAV only the ego-state, meanwhile refining the global extrinsic transformations w.r.t. (with respect to) the teammates. The ego-state and global extrinsic transformations are estimated efficiently within an Error State Iterated Kalman Filter (ESIKF) framework, by tightly fusing LiDAR point-cloud measurements, IMU measurements, and observed teammate locations (\ie, mutual observation measurements), which are enhanced by careful measurement modeling and temporal compensation. In each step of the state estimation, we marginalize the extrinsic states of all teammate UAVs not observed in the LiDAR. This state marginalization along with a degeneration evaluation method prevents the state dimension (hence computational complexity) from growing with the swarm size, effectively enhancing the scalability of our system to larger swarms.}

This paper \textcolor{black}{is extended from} our previous work\cite{zhu2023swarm}, which proposed the general framework of swarm LiDAR-inertial odometry. Compared to \textcolor{black}{the previous work} \cite{zhu2023swarm}, \textbf{\textcolor{black}{this paper proposes five crucial extensions}}:
\begin{enumerate}\color{black}
    \item
    \textcolor{black}{Factor graph optimization for efficient teammate identification and global extrinsic calibration, which largely decreases the complexity and energy consumption of the swarm initialization.} \textcolor{black}{Specifically, the number of flights required in the initialization of a swarm with $N$ UAVs is reduced from $O(N)$ to $O(1)$.}
    \item 
    \textcolor{black}{A novel state marginalization strategy and a LiDAR degeneration evaluation method that alleviate the computational burden and to} \textcolor{black}{enhance the swarm scalability.} \textcolor{black}{The marginalization reduces the growth rate of the state estimation complexity from cubic to sub-linear.}
    \item 
    Detailed measurement modeling and carefully designed temporal compensation of the mutual observation measurements,  \textcolor{black}{to compensate for the temporal mismatch due to asynchronous sensor measurements among different UAVs.}
    \item 
    \textcolor{black}{Comprehensive simulation and real-world experiments verifying the effectiveness of Swarm-LIO2, \ie, support of large swarm scales (tested 5 UAVs in the real-world as shown in Fig.~\ref{fig:cover} and 40 UAVs in simulation), robust to degenerated scenes, and allows the dynamic change of swarm size with online joining or dropping out of any teammate UAVs.}
    \item 
    \textcolor{black}{An open source implementation of the proposed system, termed as \textbf{Swarm-LIO2}, including} \textcolor{black}{source codes of the algorithms} \textcolor{black}{and hardware designs of our aerial platforms.}
\end{enumerate}

\section{Related Works}
In this section, we first review the mainstream frameworks of the state estimation for robotic swarm systems. Then we discuss the existing swarm initialization approaches, which is the core module of swarm state estimation.
\subsection{State Estimation for Robotic Swarm}

In the past few years, multi-robot systems have flourished and some great collaborative SLAM methods for ground robotic swarm systems have been proposed to achieve robust ego-state estimation and consistent mapping \cite{chang2022lamp, tian2022kimera, maplab, lajoie2023swarm}. 
Chang \etal \cite{chang2022lamp} proposed a multi-robot SLAM system in which each robot sends its single-robot odometry result and constructed submap to a centralized base station to perform loop-closure detection and joint pose graph optimization. The wheeled robot platform used in \cite{chang2022lamp} is equipped with abundant sensors, including three Velodyne LiDARs and a complete mobile LiDAR scanner system. Similarly, in \cite{maplab}, a centralized computer that receives map information from all robots is needed to implement loop closure correction and global optimization, and the platform for dataset collecting in \cite{maplab} contains five cameras and an Ouster LiDAR. The aforementioned centralized systems are fragile to single-point-of-failure, promoting the development of decentralized methods. In \cite{lajoie2023swarm}, Lajoie \etal proposed a decentralized multi-robot SLAM method in which each robot performs the same computation with only onboard computing resources. 
In \cite{tian2022kimera, lajoie2023swarm}, compact descriptors are exchanged which could partly decrease the communication load compared to exchanging raw map information, but still, the environmental information size will rapidly grow as the travel distance and swarm scale increases. All the aforementioned methods can be categorized into environment feature-based methods, of which the obvious weakness is that they are limited to feature-rich environments and the communication bandwidth is relatively high.

Compared to ground robots, the restricted payload capacity and endurance of aerial vehicles greatly limit the weight of sensors and the quality of computation units, which further necessitates a swarm with lightweight sensor configurations, efficient algorithms, and low-bandwidth communication. To satisfy these requirements, some aerial swarm systems \cite{zhou2021ego, weinstein2018visual, lusk2020distributed, quan2022formation} directly utilize independent VIO to achieve ego-state estimation which gives up the inter-UAV data fusion. Although the independent VIO is easy to adopt, the state estimation results suffer from inevitable drift, especially after long-distance running. To mitigate the VIO drift during the online running of swarm systems, in \cite{lajoie2020door}, the inter-UAV place recognition results are exchanged among the UAVs. Likewise in \cite{zhou2022swarm, xu2020decentralized,guo2019ultra}, the UWB module is incorporated to provide distance constraints as a complementary sensor of the camera. Apart from map and descriptor exchange mentioned above, mutual observation is another important unique feature of swarm systems, which is a lightweight type of information, avoiding large communication and computation load.
In \cite{xu2022omni,xu2022d,nguyen2020vision}, learning-based methods like YOLOv3-tiny are utilized to detect other robots to provide mutual observation measurements for VIO drift compensation.
These vision-based methods usually struggle in low-visibility environments and may fail to provide accurate 3D observation results due to imprecise distance estimation, which narrows down their applications in practical cases, \eg~in large-scale outdoor environments.

By contrast, LiDAR can provide accurate and long-range depth measurements, bringing many new opportunities for swarm state estimation. In \cite{dube2017online, zhou2021online, denniston2022loop,huang2021disco}, 3D LiDAR-based place recognition (loop closure) is widely utilized to improve state estimation accuracy. However, the large communication bandwidth greatly limits the scalability of the swarm systems.
Wasik \etal \cite{wasik2016lidar} propose a laser-based multi-robot system, laser range finders are used for each robot to estimate the distances and angles to other robots. However, the adopted 2D LiDARs do not apply to UAVs considered in this paper which fly in 3D spaces. Pritzl \etal \cite{pritzl2023fusion} utilize the LiDAR observation measurements to mitigate the VIO drift under the framework of non-linear least square optimization.

Compared to the centralized methods \cite{chang2022lamp, maplab,dube2017online,denniston2022loop}, our system is fully decentralized which would suffer no single-point-of-failure issue. 
Different from the environment feature-based methods \cite{chang2022lamp,lajoie2023swarm,tian2022kimera,lajoie2020door,dube2017online,huang2021disco}, our approach fuses the mutual observation measurements under ESIKF framework, leading to quite low communication bandwidth and efficient computation. Compared to the camera-based \cite{zhou2021ego,zhou2022swarm,xu2022omni,nguyen2020vision} or 2D LiDAR-based methods \cite{wasik2016lidar}, our method utilizes 3D LiDAR sensor due to its capability of providing accurate and long-range depth measurements with large field of view.

\subsection{Swarm Initialization}
The critical parts of the initialization of a robotic swarm system typically contain robot detection, robot identification, and global extrinsic calibration.

\textbf{Robot detection}. Learning-based robot detection methods are widely employed for visual-based swarm systems. For instance, Nguyen \etal \cite{nguyen2020vision} propose a visual-inertial multi-UAV localization system, in which MAVNet is used to detect other teammate robots. Xu \etal \cite{xu2020decentralized} propose a visual-inertial-UWB mutual state estimation system utilizing YOLOv3-tiny for teammate robot detection. These learning-based detection approaches usually need preliminary network training, resulting in extra time and computation consumption. In \cite{saska2017system}, each robot is equipped with a circle marker for easy detection. While for the LiDAR-based swarm systems, robot detection is more difficult since LiDAR cannot provide texture and color information. In \cite{pritzl2023fusion, vrba2023onboard}, a local occupancy map is constructed and ray-casting is utilized to detect the dynamic obstacles, which is memory-intensive and time-consuming. In a similar way to \cite{saska2017system}, in our previous work \cite{zhu2023swarm}, reflective tapes are attached to each robot (UAV) and leverage the reflectivity measured by LiDAR sensors to detect teammate robot. This detection method is simple but effective, avoiding cumbersome network training. 

\textbf{Identification and global extrinsic calibration}. In a fully decentralized system, under general circumstances, each UAV estimates the states in its own global frame. Thus, After detecting objects that might be teammate robots, each robot needs to identify other teammates and calibrate the corresponding global extrinsic transformations.
In the existing literature, the global extrinsic is usually calibrated offline such as by measuring the distance between the UAVs \cite{zhou2021ego,zhou2022swarm}, resulting in quite coarse global extrinsic values. As the flight distance increases, even small deviations in the global extrinsic parameters can lead to significant drift. In \cite{nguyen2020vision,xu2020decentralized,xu2022omni}, the global extrinsic parameters are estimated online but high-quality initial estimations are necessary. Tian \etal \cite{tian2022kimera} calibrates the global extrinsic transformations by constructing a truncated least square problem that employs the inter-robot loop closure results and the odometric estimates. This method requires environmental information exchange, leading to a large communication burden.
Chang \etal \cite{chang2022lamp} place three reflective tapes (fiducial markers) with known 3D coordinates in the take-off environment. Then the LiDAR sensor on each robot would segment the three markers and thereby compute the global extrinsic by minimizing the distance between all triplets of marker positions. Compared to \cite{chang2022lamp}, we attach reflective tapes to the UAVs instead of the environment, which supports initialization outside the take-off area. Different from the one-off calibration \cite{zhou2021ego,zhou2022swarm} or the online estimation relying on good initial values \cite{nguyen2020vision,xu2020decentralized,xu2022omni}, we utilize the trajectory matching proposed in our previous work \cite{zhu2023swarm} and factor graph optimization to calibrate the global extrinsic parameters, and constantly refine them in the subsequent swarm state estimation. The whole process is fully autonomous and no initial value is required.

\section{System Overview}\label{ch:system-overview}
In this section, we outline the structure of Swarm-LIO2 and give a brief overview of its modules. Aiming to assist understanding of the proposed system, we define some important notations in Table~\ref{notations}. 
\textcolor{black}{We use $\circ$ to compactly represent the rigid transformation of a point $\mathbf p \in \mathbb R^{3\times 1}$ with pose $\mathbf T = (\mathbf R, \mathbf t) \in SE(3)$ as $\mathbf T \circ \mathbf p  \triangleq \mathbf {Rp} + \mathbf t$.}


\begin{table}[tbp]
	\renewcommand\arraystretch{1.2}
	\caption{Some Important Notations}
	\begin{center}
		\scalebox{0.85}{
			\begin{tabular}{l p{7cm} }
			\toprule
			\textbf{Notation} & \textbf{Explanation} \\ \hline
			$\boxplus/\boxminus$ &The encapsulated ``boxplus" and ``boxminus" operations on the state manifold. \\
			$t_{i,k}$ &Timestamp of the $k$-th LiDAR measurement of UAV $i$. \\ 
			$\mathbf x_{i,k},\widehat {\mathbf x}_{i,k}, \bar{\mathbf x}_{i,k}$ & The ground-true, predicted, updated state of UAV $i$ at timestamp $t_{i,k}$.\\
            $\Tilde{\mathbf x}_{i,k}$ & Error between ground-true state and its estimation at timestamp $t_{i,k}$.\\
                $\breve{\mathbf x}_{j,k}$ & The measurements utilized by UAV $i$.\\ 
			${^{G_i}}\mathbf T_{b_i}, {^{G_i}}\mathbf v_{b_i}$ & The pose (including attitude and position) and linear velocity of UAV $i$ in its own global frame $G_i$.\\
			$\mathbf b_{g_i}, \mathbf b_{a_i}, {^{G_i}}\mathbf g$ & The gyroscope bias, the accelerometer bias, and the gravity vector of UAV $i$. \\
                $\mathbf{P}_i$ & The covariance of UAV $i$'s state.\\
                ${^i}\tau_j$ &The temporal offset between UAV $j$ and UAV $i$.\\
                ${^{G_i}}\mathbf T_{G_j}$ & The extrinsic transformation from the global reference frame of UAV $j$ to that of UAV $i$, consisting of rotation ${^{G_i}}\mathbf R_{G_j}$ and translation ${^{G_i}}\mathbf p_{G_j}$. \\
			${^{G_i}}\mathbf x_{b_j}$ & The state of UAV $j$ estimated by UAV $i$.\\
                ${^{b_i}}\breve{\mathbf p}_{b_j}$ & The relative position of UAV $j$ with respect to UAV $i$ measured in the latter and hence represented by the latter's body frame, also referred to as active mutual observation. \\
                $N$ & The total number of the UAVs in the swarm.\\
                \bottomrule
		    \end{tabular}
		    }
	\end{center}
	\label{notations}
  \vspace{-3mm}
\end{table}

\begin{figure}[t]
	\centering 
	\includegraphics[width=0.9\linewidth]{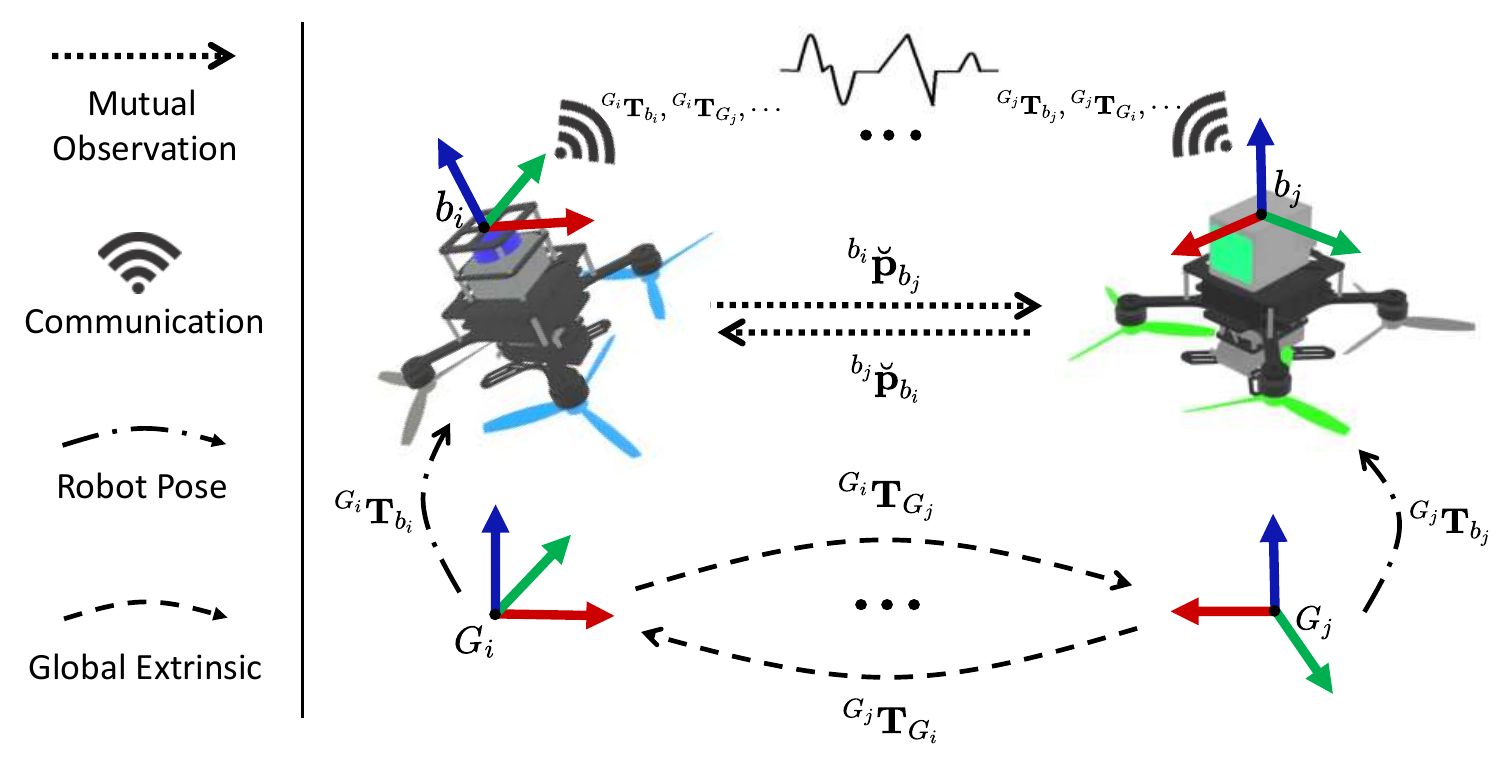}
	\caption{The illustration of the swarm state estimation problem.}
	\label{fig:big_picture}
  \vspace{-3mm}
\end{figure}

Consider an aerial swarm system consisting of $N$ UAVs and each one is equipped with a LiDAR and an inertial measurement unit (IMU). To achieve decentralized swarm state estimation, each UAV is required to detect, automatically, all the teammate UAVs in the system and estimate, in real-time, the state of itself (\eg, ego-state estimation) as well as of all other teammates (\eg, mutual state estimation). Performing ego-state and mutual state estimation altogether in one individual UAV is challenging due to the limited onboard computation resources and the high system dimension. Therefore, Swarm-LIO2 estimates the ego-state on each UAV and broadcasts the ego-state among the teammates. Since the ego-state is performed in each UAV's own global reference frame \textcolor{black}{(\ie, the first IMU frame)}, the extrinsic transformations among all UAV pairs' global frames also need to be calibrated. With the calibrated global extrinsic, the received teammates' ego-state can be projected to the self global reference frame, hence achieving the mutual state estimation (see Fig.~\ref{fig:big_picture}).

Summarizing the above analysis, Swarm-LIO2 has two key modules. The first module is online initialization, in which each UAV detects all the teammates and performs temporal and spatial calibration with the detected teammate. Let $i$ denote the self UAV and $j$ denote the detected teammate candidate, since the computer clocks of different UAVs are usually asynchronous, the temporal offset ${^i}\tau_j$ between the clocks of any two UAVs $i,j$ should be calibrated, which is essential for the \textcolor{black}{inter-UAV} data fusion. Then, the self UAV needs to validate the teammate identity (UAV ID, \textcolor{black}{which is a unique number assigned to each UAV once manufactured}) and, if successful, calibrate the global extrinsic transformation ${^{G_i}}\mathbf T_{G_j} = ({^{G_i}}\mathbf R_{G_j},{^{G_i}}\mathbf p_{G_j}) \in SE(3)$ w.r.t. it.

The second module is state estimation, aiming to estimate in real-time the ego-state of each UAV (\eg, pose, velocity), by fusing the self-LiDAR and IMU data as well as mutual observation measurements. When the estimated ego-state is projected to the teammates' global frames using the global extrinsic transformation, a small extrinsic error occurs in the initialization stage could lead to a large mutual state estimation error if the UAV's travel distance is long. To mitigate this error, Swarm-LIO2 refines the global extrinsic transformations online along with the ego-state.


\begin{figure*}[t]
	\centering 
	\includegraphics[width=0.85\textwidth]{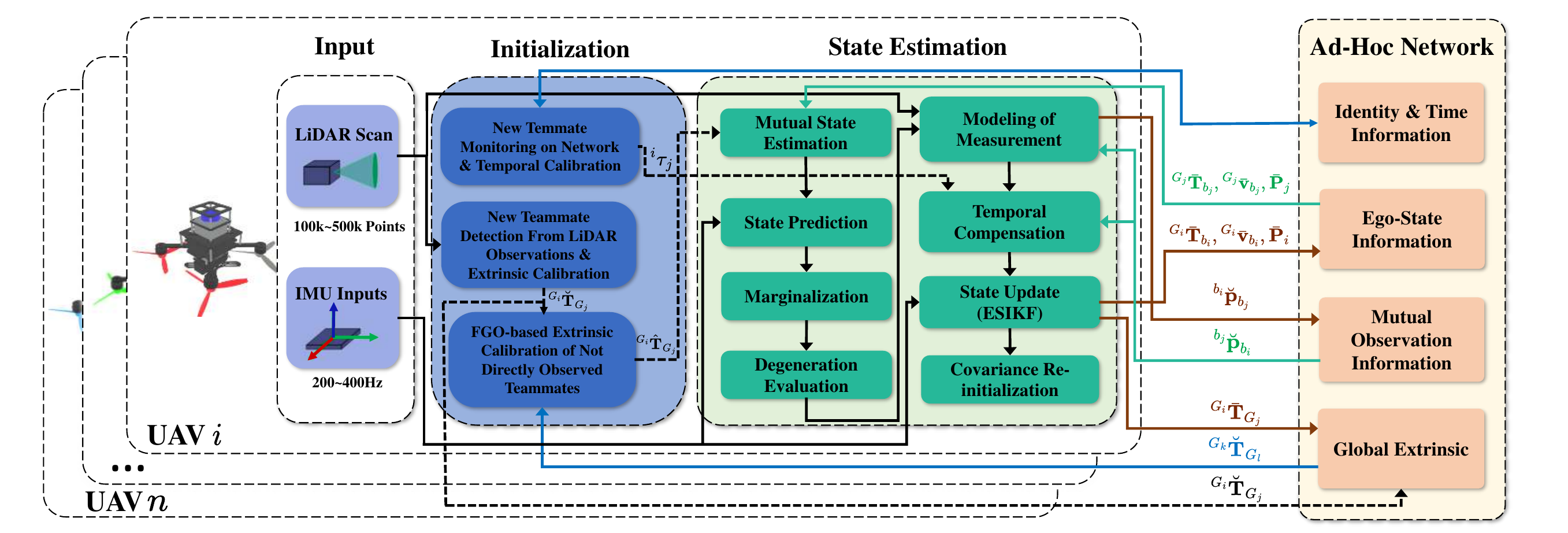}
	\caption{Framework of the proposed state estimation system for aerial swarm systems. \textcolor{black}{The dashed arrow lines mean the messages are sent only once, while the solid arrow lines mean the messages are sent constantly at the scan rate.}}
	\label{fig:framework}
  \vspace{-3mm}
\end{figure*}

The two modules of Swarm-LIO2 run in parallel on each UAV of the swarm system, as detailed in Fig.~\ref{fig:framework}. For the initialization module (Section \ref{section:initialization}), it further contains three sub-modules that run concurrently. The first sub-module monitors \textcolor{black}{new} teammate UAVs on the network and calibrates the temporal offsets ${^i}\tau_j$ w.r.t. it (Section \ref{section:sync}). The second sub-module detects \textcolor{black}{new} teammates observed in LiDAR point-cloud, and calibrates the global extrinsic transformation ${^{G_i}}\mathbf T_{G_j}$ w.r.t. it (Section \ref{section:detection}). \textcolor{black}{The calibrated global extrinsic are then sent to the third sub-module of the self-UAV and teammate UAVs on the network. Then, the third sub-module receives the global extrinsic from the second sub-module of the self-UAV or teammate UAVs on the network, based on which the global extrinsic w.r.t. teammates not observed in LiDAR point-cloud are calibrated} via a factor graph optimization (Section \ref{section:FGO}). Once the global extrinsic ${^{G_i}}\mathbf T_{G_j}$ w.r.t. UAV $j$ is calibrated, UAV $j$ is considered as a valid teammate whose state will be added to and estimated in the state estimation module. \textcolor{black}{Meanwhile, the extrinsic ${^{G_i}}\mathbf T_{G_j}$ is sent to the state estimation module for further refinement}. 

\textcolor{black}{For the state estimation module (Section \ref{section:state estimation}), it estimates the swarm state, which consists of the ego-state and the global extrinsic transformations w.r.t. all teammates}. To reduce the state dimension, Swarm-LIO2 performs a marginalization step (Section \ref{section:marginalization}), followed by a degeneration evaluation to evaluate the degeneration of current LiDAR measurements indicated by $\mathcal I_i$ and to perform further marginalization (Section \ref{section:degeneration_detection}). The marginalized state is then estimated in an error-state iterative Kalman filter (ESIKF) framework \cite{he2021kalman} by performing state prediction (Section \ref{section:propagation}) and iterative update (Section \ref{section:state estimation}) after measurements modeling (Section \ref{section:meas_modeling}). The measurements contain LiDAR point-cloud and mutual observation measurements \textcolor{black}{(\ie, the teammate location observed by the self UAV, denoted by ${^{b_i}}\breve{\mathbf p}_{b_j}$, and the self-location observed by the teammate, denoted by ${^{b_j}}\breve{\mathbf p}_{b_i}$, which is received from the teammate). Mutual observations have temporal mismatch,} which are temporally compensated in Section \ref{section:compensation}.
The state estimation results are finally transmitted to other teammate UAVs for their next round of state estimation (Section \ref{section:broadcast_of_state}). 
In Swarm-LIO2, all information is exchanged via a fully decentralized Ad-Hoc network \textcolor{black}{infrastructure under the IEEE 802.11 architecture (\ie, IBSS)}, which is broadly supported by commonly used WiFi modules and can be configured by programming the WiFi driver \cite{wu2004ad}.

\section{Swarm Initialization}\label{section:initialization}

In this section, we introduce the three sub-modules of the swarm initialization process \textcolor{black}{running on each UAV of the swarm system}. The first sub-module \textcolor{black}{(Section \ref{section:sync})} monitors \textcolor{black}{new teammate UAVs on the network} and calibrates the temporal offsets w.r.t. them. The second sub-module \textcolor{black}{(Section \ref{section:detection})} \textcolor{black}{monitors new teammate UAVs from LiDAR measurements}, validates their identities, and calibrates the global extrinsic transformations w.r.t. them via trajectory matching. The third sub-module \textcolor{black}{(Section \ref{section:FGO}) monitors the extrinsic updates from the second module or the network, and calibrates the global extrinsic transformations w.r.t. other teammates that are not directly observed by its LiDAR, via a novel factor graph optimization}. 

\subsection{\textcolor{black}{New Teammate Monitoring on Network and Temporal Calibration}}\label{section:sync}
\textcolor{black}{The first sub-module detects teammate UAVs on the network, maintains the connection with the found teammates, and performs temporal offset calibration w.r.t. them. To achieve this}, each UAV would continuously broadcast its identity information in the Ad-Hoc network, including its UAV ID and Internet Protocol (IP) address, at a fixed frequency of \SI{1}{Hz}. This identity information is commonly called the ``heartbeat'' \textcolor{black}{packet}, used for teammate monitoring and communication status maintenance. The ``heartbeat" \textcolor{black}{packet} could also be encrypted if necessary to prevent UAV information leakage or cyber-attacks. Upon receiving identity information from a teammate, the self-UAV adds the teammate to its teammate list and maintains the connection status continuously. For each teammate in the teammate list, the self-UAV assigns one of two states: connected or disconnected. After a teammate is added to the teammate list, its corresponding status is initialized as ``connected".
If the UAV fails to receive identity information from a teammate for \textcolor{black}{two seconds}, the teammate's state is set to ``disconnected". Upon receiving the identity information from the disconnected teammate again, the state is switched back to ``connected".

After discovering a new teammate \textcolor{black}{on the network}, the crucial temporal offset w.r.t. the teammate UAV is calibrated. For each teammate UAV, a decentralized temporal calibration method based on the peer-delay mechanism in Precision Time Protocol (PTP)\cite{watt2015understanding} is utilized to acquire the temporal offset corresponding to it. 
\textcolor{black}{
The self-UAV would send request messages to each teammate UAV and receive response messages from teammates. By leveraging the timestamps of these messages, the self-UAV can calculate the temporal offset w.r.t. each teammate UAV following the principle of PTP \cite{watt2015understanding}.} To suppress random errors or fluctuations, this process is repeated 30 times, and the average value of ${^i}\tau_j$ is adopted. \textcolor{black}{Since the clock drift among different UAVs is negligible within the typical UAV flight time (\eg, less than an hour), estimating the temporal offset one time is sufficient for actual swarm tasks, \ie, for any UAV $i$, once its temporal offset w.r.t. UAV $j$, ${^i}\tau_j$, is obtained, the corresponding temporal calibration is considered as complete and no request-response messages will be communicated with UAV $j$. If the clock drift is significant, the temporal offset can be estimated constantly at a fixed frequency, \eg, \SI{1}{Hz}.}
This mechanism is performed for both UAVs in each pair of the swarm system and is robust to single-point-of-failure due to the absence of a designated master clock. Each UAV performs temporal offset calibration with every teammate UAV newly found in the network. The calibrated temporal offset ${^i}t_{j}$ is stored in a Hash table where the key is UAV ID and the value is temporal offset.
When a UAV receives any data from a teammate, the data is stamped with the teammate's clock. To use the data for the self-UAV, the received data will have its timestamp modified according to the temporal offset obtained by the fast \textcolor{black}{and efficient} lookup of the Hash table.

\subsection{\textcolor{black}{New Teammate Detection from LiDAR Observations and Extrinsic Calibration}}\label{section:detection}
For any teammate in the teammate list, apart from calibrating the temporal offset, each UAV also needs to calibrate the spatial offset, \ie, the extrinsic transformations between the two UAVs' global reference frames. In this section, we calibrate the global extrinsic w.r.t. those teammates \textcolor{black}{observed by the LiDAR on the self-UAV}. 

We propose a novel reflectivity filtering and cluster extraction-based teammate detection method to easily detect the observed teammate UAVs from LiDAR point-cloud measurements. After accumulating the trajectory of the directly observed objects over a certain time, a trajectory matching-based identification and global extrinsic calibration method are used for fast swarm initialization.

 \begin{algorithm}[t]
 \SetAlgoLined
  \footnotesize
  \caption{\textcolor{black}{New Teammate Detection,} Identification, and Global Extrinsic Calibration}
  \label{alg:initialization}
  \KwIn{
    A scan of LiDAR raw points \textcolor{black}{excluding points on known teammate UAVs} ${^{b_i}} \mathcal P$, UAV $i$'s odometry ${^{G_i}}\mathbf T_{b_i}$, time interval of LiDAR input $\Delta t$, UAV $j$'s position trajectory $^{G_j}\mathcal T_j$, threshold of trajectory matching residual $thr$
  }
  \KwOut{
    \textcolor{black}{The number of clusters $M$,} Global extrinsic transformation ${^{G_i}}\mathbf T_{G_j}$, tracked position of UAV $j$ in UAV $i$'s global frame ${^{G_i}} \mathbf p_{b_j}$
  }  
  \BlankLine
  ${^{b_i}} \mathcal P_h$ = $\mathtt{ReflectivityFiltering}$(${^{b_i}} \mathcal P$)\; \label{alg:ReflectivityFiltering}
  $\textcolor{black}{M,} {^{b_i}} \breve{\mathbf p}_m$ = $\mathtt{FastEuclideanClustering}$(${^{b_i}} \mathcal P_h$)\; \label{alg:EuclideanClustering}
  \For{$m = 1:M$}{
    ${^{G_i}} \bar{\mathbf p}_m$ = $\mathtt{TemporaryTracking}$($\Delta t, {^{G_i}}\mathbf T_{b_i}, {^{b_i}} \mathcal P, {^{b_i}} \breve{\mathbf p}_m$)\; \label{alg:TemporaryTracking}
    $^{G_i}\mathcal T_m$.PushBack(${^{G_i}} \bar{\mathbf p}_m$)\; \label{alg:TrajAcc}

    \If(){$\mathtt{TrajExcited}$($^{G_i}\mathcal T_m$) \label{alg:TrajExcited}}
    {
      \For{$ j = 1:N$; $j \neq i$}
      {
        $res, \mathbf T$ = $\mathtt{TrajMatching}$(${^{G_j}}\mathcal T_j, {^{G_i}}\mathcal T_m$)\; \label{alg:TrajMatching}
        \If(){$res < thr$}
        {
          ${^{G_i}}\mathbf T_{G_j} = \mathbf T$ \;
          ${^{G_i}} \mathbf p_{b_j}$ = ${^{G_i}} {\mathbf p}_m$\;
          break\;
        }
      }
    }
  }
\end{algorithm}

To implement the above method, for each UAV, several reflective tapes are attached to its body, so that it can be easily detected and tracked by other teammates based on the reflectivity information measured by the LiDAR sensor. The detailed implementation of this sub-module is summarized in Alg.~\ref{alg:initialization}. \textcolor{black}{After receiving a LiDAR scan}, we first undistort the raw points following \cite{xu2021fast} \textcolor{black}{and filter out the points on teammate UAVs with whom the initialization has completed (which is explained later in Section \ref{section:meas_modeling}),}
to obtain the LiDAR points ${^{b_i}} \mathcal P$, which is represented in the current body frame. Then, points with high reflectivity values exceeding a pre-defined threshold, which can be calibrated beforehand on the reflective tapes attached to each UAV, are extracted by $\mathtt{ReflectivityFiltering}$(${^{b_i}} \mathcal P$) in Line \ref{alg:ReflectivityFiltering} of Alg.~\ref{alg:initialization}. Then in Line \ref{alg:EuclideanClustering}, the high-reflectivity points ${^{b_i}} \mathcal P_h$ are efficiently clustered by $\mathtt{FastEuclideanClustering}$ (FEC) \cite{cao2022fec}, which aims to detect new potential teammate UAVs. 
Each detected object, with clustered centroid ${^{b_i}} \breve{\mathbf p}_m$, is then tracked by a Kalman filter-based temporary tracker \textcolor{black}{based on the assumption of constant velocity in Line \ref{alg:TemporaryTracking} \cite{zhu2023swarm}}.

The tracked trajectory of each potential new teammate is accumulated (Line \ref{alg:TrajAcc}) for subsequent identification and global extrinsic calibration. To achieve this, all UAVs in the swarm system will exchange their estimated ego-states (in their own global frames) with others. The ego-state of each UAV is estimated by \textcolor{black}{the state estimation module (Section \ref{section:state estimation}), which runs in parallel to the initialization module as explained in Section \ref{ch:system-overview}, based on LiDAR points, IMU data, and mutual observations of teammate UAVs if available.} 

Let ${^{G_i}}{\mathcal T}_{m} = \{{^{G_i}}\bar{\mathbf p}_{m,\kappa}, \kappa = 1,\cdots, \mathcal K\}$ denote the trajectory of the $m$-th temporary tracker and ${^{G_j}}{\mathcal T}_{j} = \{{^{G_j}} \breve{\mathbf p}_{b_j,\kappa}, \kappa = 1, \cdots, \mathcal K\}$ represent the trajectory received from UAV $j$, the $m$-th tracked object is identified as UAV $j$ if the following trajectory matching problem has unique optimal solution with a residual below a certain threshold:

\begin{small}
    \begin{equation}\label{trajmatching}
    \arg\min_{^{G_i}\mathbf T_{G_j}}\sum_{\kappa = 1}^\mathcal K \dfrac{1}{2}\| {^{G_i}}\bar{\mathbf p}_{m,\kappa} - \textcolor{black}{{^{G_i}}\mathbf T_{G_j} \! \circ \! {^{G_j}}\breve{\mathbf p}_{b_j,\kappa}}  \|,
\end{equation}
\end{small}

Considering possible short-term communication disconnection, some data of ${^{G_j}} \breve{\mathbf p}_{b_j}$ might be lost. Thus, we only pick ${^{G_i}}\bar{\mathbf p}_{m,\kappa}$ that has close timestamp with ${^{G_j}} \breve{\mathbf p}_{b_j,\kappa}$ to participate in trajectory matching. Besides, to avoid large computing time due to too much data, we use a sliding window of the most recent $\mathcal K$ positions for matching. \textcolor{black}{We implement the above trajectory matching process as a function $\mathtt{TrajMatching}$(${^{G_j}}\mathcal T_j, {^{G_i}}\mathcal T_m$) in Line \ref{alg:TrajMatching}.}

\begin{figure*}[t]
    \setlength{\abovecaptionskip}{-0.1cm}
	\centering 
	\includegraphics[width=0.95\textwidth]{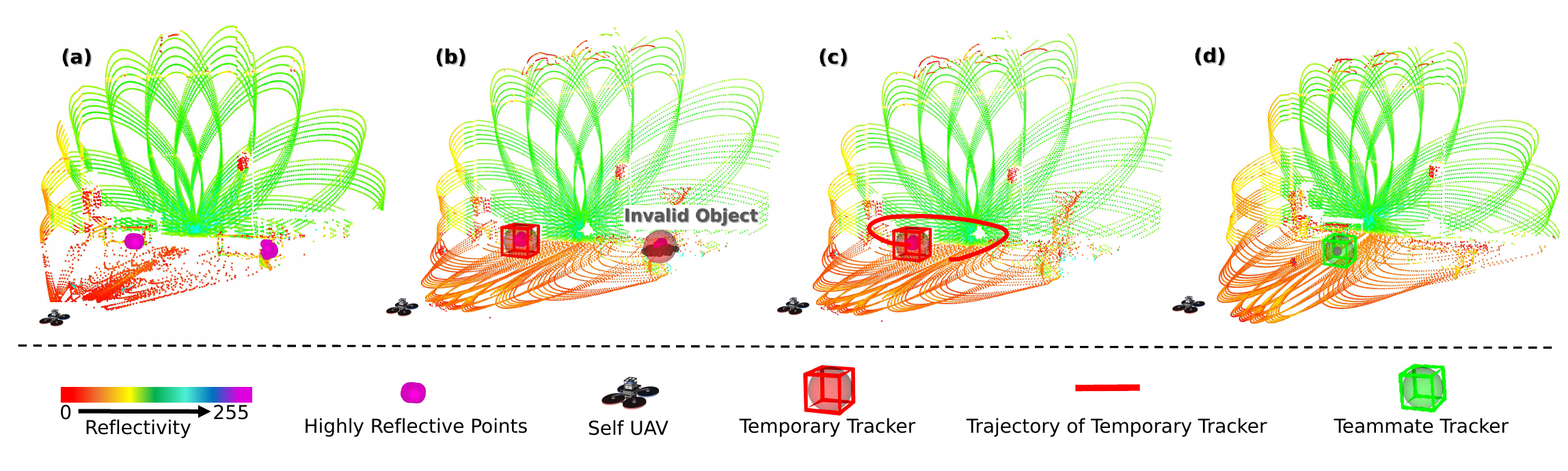}
	\caption{Illustration of the initialization for newly detected objects, point-cloud is colored by reflectivity. Here the self-UAV needs to detect and identify other teammate UAVs in its FoV. The sphere represents the predicted region and the center of the bounding box represents the updated position of the tracker. (a) Reflectivity filtering. (b) Outlier rejection by discarding objects with too large size. (c) Track real potential teammates and accumulate the trajectory. (d) After trajectory matching, the object is identified as a teammate UAV with a correct UAV ID.}
	\label{fig:initialization}
 \vspace{-3mm}
\end{figure*}

Since no unique transformation can be determined from (\ref{trajmatching}) if the involved trajectories are straight lines \cite{PointRegister}, the trajectories of those tracked objects are constantly evaluated by $\mathtt{TrajExcited}$($^{G_i}\mathcal T_m$) in Line \ref{alg:TrajExcited}. 
 \textcolor{black}{Once the condition of $\mathtt{TrajExcited}$($^{G_i}\mathcal T_m$) is satisfied, the trajectory matching will be performed.}
Let ${^{G_i}} \bar {\mathbf p}_{m}^c$ represent the centroid of $^{G_i}\mathcal T_m$, $\mathtt{TrajExcited}$($^{G_i}\mathcal T_m$) assesses the excitation (shape) of $^{G_i}\mathcal T_m$ by computing the singular values of matrix $\mathcal H \in \mathbb R^{3\times 3}$:
\begin{equation}
    \mathcal H \triangleq \sum_{\kappa=1}^\mathcal K ({^{G_i}} \bar{\mathbf p}_{m,\kappa} - {^{G_i}} \bar{\mathbf p}_{m}^c)\cdot ({{^{G_i}} \bar{\mathbf p}_{m,\kappa}} - {^{G_i}} \bar{\mathbf p}_{m}^c)^T.
\end{equation}

If the second largest singular value is larger than a given threshold, \textcolor{black}{it means the positions of the trajectory $^{G_i}\mathcal T_m$ do not lie on a straight line, which ensures a unique solution in} $\mathtt{TrajMatching}$(${^{G_j}}\mathcal T_j, {^{G_i}}\mathcal T_m$) \cite{PointRegister}. The matching for temporary tracker trajectory ${^{G_i}}\mathcal T_m$ is performed with all received teammate UAV's trajectory ${^{G_j}}\mathcal T_j, j = 1,\cdots, N$ until the matching error is smaller than a given threshold, indicating that the object $m$ is essentially the observation of teammate with UAV ID $j$, and the solution of \eqref{trajmatching} gives an initial estimation of the global extrinsic ${^{G_i}}\breve{\mathbf T}_{G_j}$. \textcolor{black}{Meanwhile, the $m$-th temporary tracker will be removed from the temporary tracker list and become a teammate tracker that will used in the state estimation module (Section \ref{section:state estimation})}. An example of the trajectory matching-based initialization pipeline is illustrated in Fig.~\ref{fig:initialization}. 

\textcolor{black}{Finally, the extrinsic obtained by trajectory matching ${^{G_i}}\breve{\mathbf T}_{G_j}$ is sent to the third sub-module of the self-UAV as well as all teammate UAVs on the teammate list. Note that the sending of ${^{G_i}}\breve{\mathbf T}_{G_j}$ occurs only once since after the temporary tracker has been removed, no trajectory matching will be performed to produce the extrinsic ${^{G_i}}\breve{\mathbf T}_{G_j}$ in the next cycle. }

\subsection{Factor Graph Optimization-based Global Extrinsic Calibration of Not Directly Observed Teammates}\label{section:FGO}

\begin{figure}[tbp]
	\setlength\abovecaptionskip{-0.05\baselineskip}
	\centering
	\includegraphics[width=0.8\linewidth]{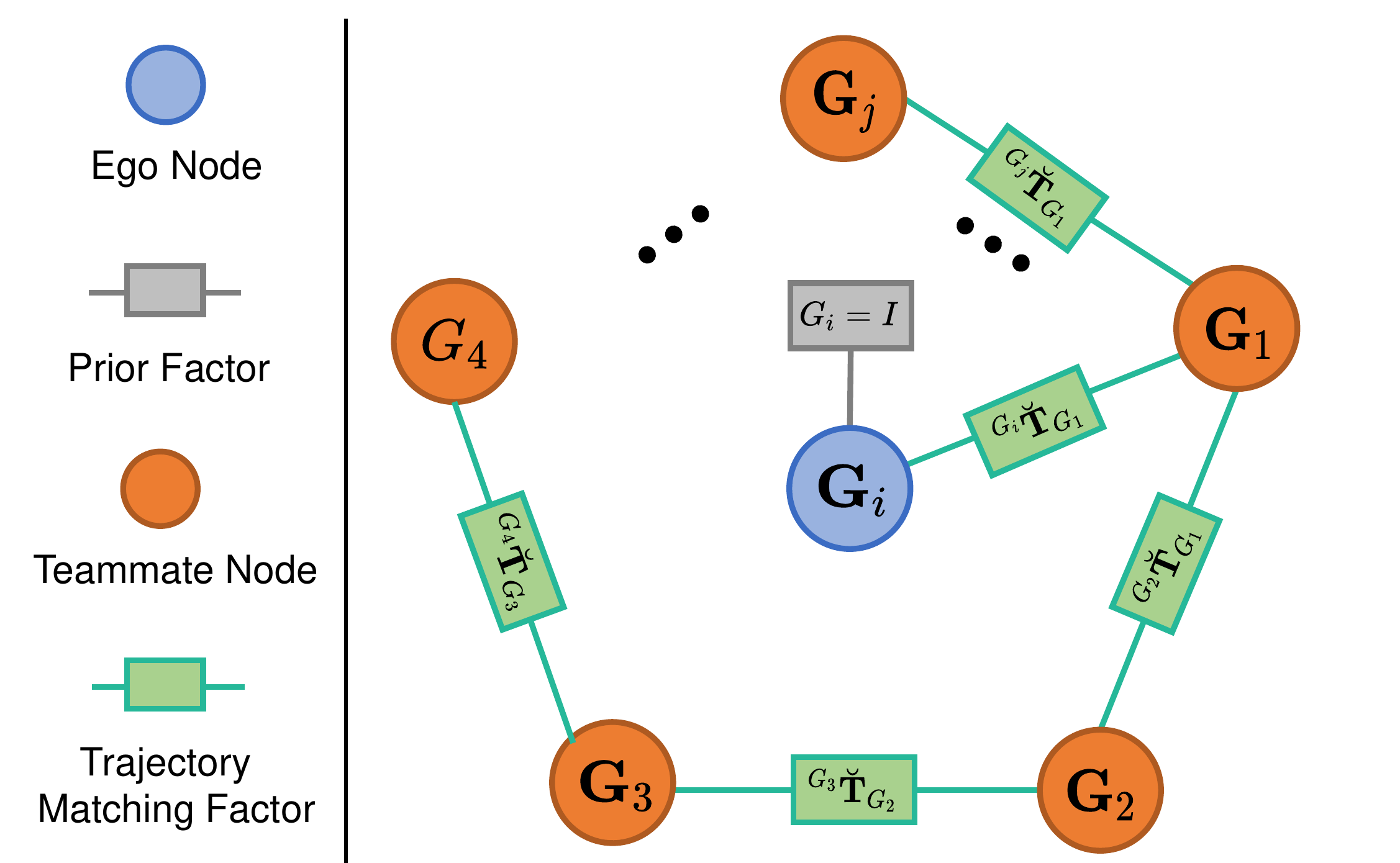}
	\caption{The illustration of the decentralized factor graph optimization-based global extrinsic calibration. To handle the gauge freedom of the factor graph, we insert a prior factor as $G_i = I$. Note that the factors $^{G_i}\breve{\mathbf T}_{G_j}$ are global extrinsic transformations received from teammate UAVs or obtained by direct trajectory matching on the self-UAV.} 
	\label{fig:factor_graph} 
 \vspace{-3mm}
\end{figure} 

Apart from the trajectory matching-based identification method detailed above, which was originally presented in our previous work \cite{zhu2023swarm}, a novel decentralized factor graph optimization method is proposed to \textcolor{black}{calibrate the global extrinsic transformations w.r.t. not directly observed teammates,} which expedites the identification and the swarm initialization.
In Swarm-LIO2, each UAV will share the global extrinsic transformations obtained via trajectory matching with all teammate UAVs in the teammate list. Then, each UAV constructs and maintains a factor graph (see Fig.~\ref{fig:factor_graph}) where the variables $G_i,G_j,\cdots$ are the global reference frames of all UAVs (including teammates with or without direct observation) and the factors $^{G_i}\breve{\mathbf T}_{G_j}$ are the global extrinsic transformation between any two UAVs, which could be calibrated by the self-UAV using the trajectory matching or received from teammate UAVs. By fixing the global frame $G_i$ of the self-UAV, it can use the global extrinsic transformations $^{G_i}\breve{\mathbf T}_{G_j}$ as constraints to solve for the global frames of all other UAVs who are connected to the self-UAV in the factor graph. Subsequently, the global extrinsic between each UAV (with and without direct observations) and the ego-UAV can be deduced as $^{G_i}\hat{\mathbf T}_{G_j} = G_iG_j^{-1}$.

\textcolor{black}{The third sub-module runs after the second sub-module, hence running recurrently at the scan rate too. Specifically, it receives extrinsic $^{G_i}\breve{\mathbf T}_{G_j}$ from the second sub-module and extrinsic $^{G_k}\breve{\mathbf T}_{G_l}$ from the network. If the received extrinsic (either from the second sub-module or from the network) corresponds to a new edge that did not exist before in the factor graph, an edge corresponding to this extrinsic will be created in the factor graph. Otherwise, the received extrinsic will be dumped to avoid information reuse. On the other hand, if there are multiple global extrinsic transformations on the same edge, such as the $^{G_i}\breve{\mathbf T}_{G_j}$, which is obtained through trajectory matching on the self-UAV $i$, and $^{G_j}\breve{\mathbf T}_{G_i}$, which is obtained through trajectory matching on the teammate UAV (received on the network)}, the average of these global extrinsic transformations is computed and used as a factor, \textcolor{black}{which can effectively save the number of factors in the factor graph}. In case the factor graph is updated, an optimization process is performed using iSAM2\cite{kaess2012isam2}, and the optimized global extrinsic, if not sent before, is sent to the state estimation module \textcolor{black}{as the initial estimation $^{G_i}\hat {\mathbf T}_{G_j}$ for the online global extrinsic refinement. }

\textcolor{black}{
\textbf{Remark 1: } As can be seen, the first sub-module runs concurrently with the second and the third sub-modules at different frequencies. The first sub-module runs at 1 Hz, while the second and the third sub-modules run at the scan rate. The recurrent nature of the three sub-modules allows the swarm to discover, identify, and calibrate the global extrinsic w.r.t. new joining UAVs in the middle of a mission. }

\textcolor{black}{\textbf{Remark 2: } For each of the three sub-modules, it breaks into two parts. The first parts of the three sub-modules monitor for new teammates on the network, new teammates observed by LiDAR measurements, or new edges in the factor graph, respectively. The second parts conduct the temporal calibration, trajectory matching, and factor graph optimization, respectively.  While the first parts run at their respective frequencies, the second parts run only when the first parts detect new teammates or edges. 
}

\section{Decentralized Swarm State Estimation}\label{section:state estimation}
In this section, we will introduce the fully decentralized swarm state estimation module, including mutual state estimation, ESIKF-based ego-state estimation, and global extrinsic refinement.

\subsection{Mutual State Estimation} \label{section:mutual_state_estimation}

One crucial mission of each UAV in a swarm system is to estimate any other UAV $j$'s state ${^{G_i}}\mathbf x_{b_j}$ (including pose $^{G_i}\mathbf T_{b_j}$ and velocity $^{G_i}\mathbf v_{b_j}$) in UAV $i$'s global frame, called mutual state estimation. This is significant for various swarm applications, such as mutual collision avoidance, formation flight, etc. \textcolor{black}{However, estimating the full state ${^{G_i}}\mathbf x_{b_j}$ of all the teammate UAVs are high-dimensional tasks that are computationally demanding. To reduce the system complexity, we propose to estimate the ego-state only on each UAV, denoted by $^{G_i}\bar{\mathbf x}_{b_i}$. The estimated ego-states are exchanged through the network. Then, a UAV $i$ can estimate a teammate $j$'s state, denoted by ${^{G_i}} \bar{\mathbf x}_{b_j}$, } by directly projecting the received UAV $j$'s ego-state into UAV $i$'s global frame using the global extrinsic transformation ${^{G_i}}{\mathbf T}_{G_j} = ({^{G_i}}{\mathbf R}_{G_j}, {^{G_i}}{\mathbf p}_{G_j})$:

\begin{small}
    \begin{equation}\label{eq:mutual-state-est}
        \begin{aligned}
            ^{G_i}\bar{\mathbf T}_{b_j}  &= {^{G_i}}{\mathbf T}_{G_j} {^{G_j}}\breve{\mathbf T}_{b_j},\\
            {^{G_i}}\bar{\mathbf v}_{b_j}  &= \textcolor{black}{{^{G_i}}{\mathbf R}_{G_j}} {^{G_j}}\breve{\mathbf v}_{b_j}.
        \end{aligned}
    \end{equation}
\end{small}
\textcolor{black}{where ${^{G_j}}\breve{\mathbf T}_{b_j}$ is the received ego-state of UAV $j$. Note that ${^{G_j}}\breve{\mathbf T}_{b_j}$ is denoted as ${^{G_j}}\bar{\mathbf T}_{b_j}$ on UAV $j$ (since it is an estimation on UAV $j$), but has an accent $\breve{(\cdot)}$ on UAV $i$ (since it acts as a measurement for UAV $i$). }

\textcolor{black}{A problem in the above process is that it requires knowing the global extrinsic transformation ${^{G_i}}{\mathbf T}_{G_j} = ({^{G_i}}{\mathbf R}_{G_j}, {^{G_i}}{\mathbf p}_{G_j})$. Although they can be calibrated by the initialization process described in Section \ref{section:initialization}, possible errors could still remain. We propose to continually refine these extrinsic transformations along with the ego-state estimation in the state estimation module. Denote ${^{G_i}}\bar{\mathbf T}_{G_j} = ({^{G_i}}\bar{\mathbf R}_{G_j}, {^{G_i}}\bar{\mathbf p}_{G_j})$ the refined extrinsic transformation, the mutual state of teammate $j$ can be computed by (\ref{eq:mutual-state-est}) with ${^{G_i}}{\mathbf T}_{G_j} = ({^{G_i}}{\mathbf R}_{G_j}, {^{G_i}}{\mathbf p}_{G_j})$ being replaced by ${^{G_i}}\bar{\mathbf T}_{G_j} = ({^{G_i}}\bar{\mathbf R}_{G_j}, {^{G_i}}\bar{\mathbf p}_{G_j})$}. This mechanism enables smooth and stable mutual state estimation even in situations where frequent mutual observation losses occur due to occlusions or teammates entering and exiting the field of view (FoV).

\textcolor{black}{To refine the extrinsic transformation in the state estimation, for each new teammate $j$ whose extrinsic was calibrated in the initialization module (Section \ref{section:initialization}), we append ${^{G_i}}{\mathbf T}_{G_j} = ({^{G_i}}{\mathbf R}_{G_j}, {^{G_i}}{\mathbf p}_{G_j})$ to the existing state vector of UAV $i$, so its value can be estimated along with other states in a unified ESIKF framework. Moreover, the calibrated extrinsic from the initialization module immediately serves as the initial estimation $^{G_i}\widehat {\mathbf T}_{G_j}$ of the appended state component ${^{G_i}}{\mathbf T}_{G_j}$}.

\subsection{State and Covariance Prediction}\label{section:propagation}
We first introduce the scheme of the state and covariance prediction. \textcolor{black}{For illustration, we select UAV $i$ as the self-UAV and assume $N \!\! - \!\! 1$ teammate UAVs have been found and calibrated in the initialization module}. Let $\tau$ denote the IMU measurement index during the $k$-th LiDAR frame, the discrete state transition model is shown below:
\begin{equation}\label{discrete_model}
    \mathbf x_{i,\tau+1} = \mathbf x_{i,\tau} \boxplus(\Delta t_\tau \mathbf f_i(\mathbf x_{i,\tau}, \mathbf u_{i,\tau}, \mathbf w_{i,\tau})),
\end{equation}
 where $\Delta t_\tau$ is the time interval between two consecutive IMU measurements, $\mathbf x_{i,\tau}$ denotes ground-truth of the state at the $\tau$-th IMU measurement of the $i$-the UAV, whose timestamp is $t_{i,\tau}$. Furthermore, we use the notation $\boxplus/\boxminus$ defined in \cite{hertzberg2013integrating} to compactly represent the ``plus" on the state manifold. Specifically, for the state manifold $SO(3) \times \mathbb R^n$ in (\ref{kinematic_model}), the $\boxplus$ operation and its inverse operation $\boxminus$ are defined as
 
 \begin{small}
     \begin{equation*}
    \begin{bmatrix}
    \mathbf R \\ \mathbf a
    \end{bmatrix} \boxplus
    \begin{bmatrix}
    \mathbf r \\ \mathbf b
    \end{bmatrix}=
    \begin{bmatrix}
    \mathbf R\text{Exp}(\mathbf r) \\ \mathbf {a+b}
    \end{bmatrix};
    \begin{bmatrix}
    \mathbf R_1 \\ \mathbf a
    \end{bmatrix} \boxminus
    \begin{bmatrix}
    \mathbf R_2 \\ \mathbf b
    \end{bmatrix}=
    \begin{bmatrix}
    \text{Log}(\mathbf R_2^T \mathbf R_1) \\ \mathbf {a-b}
    \end{bmatrix}
\end{equation*}
 \end{small}

where $\mathbf R, \mathbf R_1, \mathbf R_2 \in SO(3),\mathbf r \in \mathbb R^3 , \mathbf {a,b} \in \mathbb R^n$, $\text{Exp}(\cdot): \mathbb{R}^3 \mapsto SO(3) $ is the exponential map on $SO(3)$\cite{hertzberg2013integrating} and $\text{Log}(\cdot): SO(3) \mapsto \mathbb{R}^3$ is its inverse logarithmic map.

The state vector $\mathbf x_i$, the process noise vector $\mathbf w_i$, and the input $\mathbf u_i$, omitting the the time index, are defined as:

\begin{small}
\begin{equation}\label{kinematic_model}
\begin{aligned}
\mathbf x_i &\triangleq
[
^{G_i}\mathbf R_{b_i}^T \ \ 
^{G_i}\mathbf p_{b_i}^T\ \ 
^{G_i}\mathbf v_{b_i}^T\ \ 
\mathbf b_{g_i}^T\ \ 
\mathbf b_{a_i}^T\ \ 
^{G_i}\mathbf g^T \\
& \qquad \cdots \quad
^{G_i}\mathbf R_{G_j}^T\ \ 
^{G_i}\mathbf p_{G_j}^T\ \ 
\cdots 
]^T\in \mathcal M,\\
\mathbf w_i  &\triangleq 
\begin{bmatrix}
\mathbf n_{g_i}^T  \!&\! 
\mathbf n_{a_i}^T  \! & \! 
\mathbf  n_ {\mathbf b_{gi}}^T  \!& \!
\mathbf n_{\mathbf b_{ai}}^T
\end{bmatrix}^T,\\
\mathbf u_i  &\triangleq 
\begin{bmatrix} 
\boldsymbol{\omega}_{m_i}^T  \!& \!
\mathbf a_{m_i}^T
\end{bmatrix}^T,
\end{aligned}
\end{equation}
\end{small}
where $j = 1, 2, \cdots, i-1, i+1, \cdots, N$. 

The discrete state transition function $\mathbf f_i$ is defined as:

\begin{small}
    \begin{equation}\label{transition_func}
        \mathbf f_i  \triangleq
\begin{bmatrix}
    \boldsymbol{\omega}_{m_i}  - \mathbf b_{g_i}  -\mathbf n_{g_i} \\
^{G_i}\mathbf v_{b_i} + \frac{1}{2}(^{G_i}\mathbf R_{b_i} (\mathbf a_{m_i} \!-\!\mathbf b_{a_i} \!- \!\mathbf n_{a_i}) \!+\! {^{G_i}}\mathbf g)\Delta t_\tau\\ 
^{G_i}\mathbf R_{b_i} (\mathbf a_{m_i} \!-\!\mathbf b_{a_i} \!- \!\mathbf n_{a_i}) \!+\! {^{G_i}}\mathbf g  \quad  \quad\\
\mathbf  n_ {\mathbf b_{g i}} \\
\mathbf n_{\mathbf b_{a i}}\\
\mathbf 0_{3\times 1} \\
\vdots \\
\mathbf 0_{3\times 1}\\
\mathbf 0_{3\times 1}\\
\vdots\\
\end{bmatrix}
    \end{equation}
\end{small}

where $\boldsymbol{\omega}_{m_i}, \mathbf a_{m_i}$ represent the IMU (gyroscope and accelerometer) measurements of UAV $i$, $\mathbf n_{g_i}$ and $\mathbf n_{a_i}$ are the white noise of IMU measurements, $\mathbf b_{g_i}$ and $\mathbf b_{a_i}$ are the IMU bias modeled as the random walk process with Gaussian noises $\mathbf n_ {\mathbf b_{g i}}$ and $\mathbf n_{\mathbf b_{a i}}$.
The meaning of each element in state vector $\mathbf x_i$ is introduced in Table~\ref{notations}, the state manifold $\mathcal M$ is defined in \eqref{dim_M} and its dimension is $18+6(N-1)$.

\begin{small}
    \begin{equation}\label{dim_M}
    \mathcal M \triangleq \underbrace{SO(3) \times \mathbb R^{15}}_{\text{dim} = 18} \times \underbrace{\cdots \times  SO(3) \times \mathbb R^{3} \times \cdots}_{\text{dim} = 6 (N-1)}
\end{equation}
\end{small}

Following the state model in (\ref{discrete_model}), the state and covariance prediction is implemented under the ESIKF framework once receiving a new IMU measurement. More specifically, the state and covariance are predicted following \eqref{discrete_model} by setting the process noise $\mathbf w_{i,\tau}$ to zero. The detailed demonstration of predication can be referred to \cite{xu2021fast,xu2022fast}.

\subsection{Error State Iterative State Update}\label{section:esikf}
The update step is implemented iteratively at the end time of the new LiDAR scan at $t_{i,k}$, fusing point-cloud measurements and mutual observation measurements (if any). In the following sections, we will introduce the measurement model of the point-cloud measurements, and the novel mutual observation measurements\textcolor{black}{, which were not present in \cite{zhu2023swarm}.}

\subsubsection{Modeling of Measurements}\label{section:meas_modeling}
In the general ESIKF framework, for any measurement $\mathbf y_k$ at the $k$-th round, we can write the measurement model as
\begin{equation}\label{obs_function}
    \mathbf y_k= \mathbf h(\mathbf x_k,\mathbf v_k)
\end{equation}
where $\mathbf h(\mathbf x_k,\mathbf v_k)$ is the measurement model depending on the true state $\mathbf x_k$ and the measurement noise $\mathbf v_k$ which is assumed to be zero mean multivariate Gaussian noise. For convenience and simplification of the description, we omit the subscript $k$ in the following formulations.

Once receiving a new LiDAR scan, motion compensation will be performed to obtain the undistorted point clouds. Then the point-to-plane distance will be calculated to generate point-cloud constraints. The details of the motion compensation can be referred to \cite{xu2021fast}. The $n$-th undistorted point of the current scan projected into the body frame is denoted by $^{b_i}\mathbf p_{n}$, let $\mathbf u_n$ represent the normal vector of the corresponding plane in the global frame $G_i$, on which lies a point ${^{G_i}}\mathbf q_{n}$. Considering the LiDAR measurement noise $\mathbf n_{p, n}$ of the $n$-th point, we obtain the measurement model of the $n$-th point measurement as \cite{xu2021fast,xu2022fast}

\begin{small}
    \begin{equation}\label{eq:point_measurement}
    \mathbf{0} =
    \underbrace{ \mathbf u_n^T \textcolor{black}{({^{G_i}}\mathbf T_{b_i} \! \circ \!( {^{b_i}}\mathbf p_{n} \! +  \! \mathbf n_{p, n})}  \!  -  \!  {^{G_i}}\mathbf q_{n})}_{\mathbf h_{p,n}(\mathbf x_{i}, \mathbf n_{p, n})}
    \end{equation}
\end{small}
which defines an implicit measurement equation about the state vector $\mathbf x_i$ containing ego-pose ${^{G_i}}\mathbf T_{b_i}$. The normal vector $\mathbf u_n$ and the point $^{G_i}\mathbf q_n$ are known vectors, and $\mathbf n_{p,n}$ is point measurement noise, both can be referred to \cite{xu2021fast,xu2022fast}.

Apart from the point-cloud measurements, the mutual observation measurements are also used for state updates, \textcolor{black}{which can be obtained by the teammate tracker that evolved from the temporary tracker in the initiation module}. Specifically, with the predicted pose $ {^{G_i}}\widehat{\mathbf T}_{b_i}$ obtained in Section \ref{section:propagation}, we can acquire the predicted position of each teammate UAV $j$ described in UAV $i$'s body frame as $^{b_i}\widehat{\mathbf p}_{b_j} = \textcolor{black}{ \left({^{G_i}}\widehat{\mathbf T}_{b_i}^{-1}{^{G_i}}\bar{\mathbf T}_{G_j} \right) \! \circ \!{^{G_j}}\breve{\mathbf p}_{b_j}}$. The LiDAR points around the predicted position $^{b_i}\widehat{\mathbf p}_{b_j}$ will be \textcolor{black}{removed from the LiDAR raw points}. \textcolor{black}{The rest of the points ${^{b_i}} \mathcal P$ will be used by $\mathtt{ReflectivityFiltering}$(${^{b_i}} \mathcal P$) in the initialization module (Section \ref{section:detection}, Algorithm \ref{alg:initialization}) for new teammate detection.
Moreover, the points around the predicted teammate positions will be used for Euclidean clustering to obtain the mutual observation measurement}. If a valid object is clustered, the centroid position of the object will be regarded as the actual position of UAV $j$ observed by UAV $i$, called ``active observation measurement" for UAV $i$ w.r.t. UAV $j$ which is denoted by ${^{b_i}}\breve{\mathbf p}_{b_j}$. Each UAV would share this active observation measurement with all teammates and meanwhile receive teammates' ones via the Ad-Hoc network. The received observation measurement from UAV $j$ is referred to as ``passive observation measurements" and is denoted by ${^{b_j}}\breve{\mathbf p}_{b_i}$, representing the self-position of UAV $i$ observed by teammate $j$.

The explicit measurement model of the active observation measurement $^{b_i}\breve{\mathbf p}_{b_j}$ can be obtained by projecting UAV $j$'s ground-true position $^{G_j}\mathbf p_{b_j}$ into UAV $i$'s body frame using the ground-true global extrinsic ${^{G_i}}\mathbf T_{G_j}$ and the ground-true ego-pose ${^{G_i}}\mathbf T_{b_i} = ({^{G_i}}\mathbf R_{b_i}, {^{G_i}}\mathbf p_{b_i})$ of UAV $i$.
Further considering that the active observation may have {measurement noise $\mathbf n_{ao,ij} \sim \mathcal N(\mathbf 0,\boldsymbol{\Sigma}_{ao,ij})$} due to incomplete point measurements on the teammate UAV $j$, the model of the active observation measurement is:

\begin{small}
    \begin{equation}
        {^{b_i}}\breve{\mathbf p}_{b_j} =  
     \textcolor{black}{ \left({^{G_i}}\mathbf T_{b_i}^{-1} {^{G_i}}\mathbf T_{G_j} \right)\! \circ \! {^{G_j}}{\mathbf p}_{b_j}} \! + \! \mathbf n_{ao,ij}.
    \end{equation}
\end{small}

This measurement equation, unfortunately, involves the ground-true position $^{G_j}\mathbf p_{b_j}$ of UAV $j$, which is not a part of the state vector $\mathbf x_i$ as defined in \eqref{kinematic_model}. To fix this issue, we leverage the estimated ego-position ${^{G_j}}\breve{\mathbf p}_{b_j}$ of UAV $j$ and its covariance $\breve{\boldsymbol{\Sigma}}_{\mathbf p_j}$, both are received from UAV $j$. Then, the ground-true position of UAV $j$ is modeled as $^{G_j}\mathbf p_{b_j} = {^{G_j}}\breve{\mathbf p}_{b_j} + \mathbf n_{\mathbf p_j}$, where the noise $\mathbf n_{\mathbf p_j} \sim \mathcal N(\mathbf 0, \breve{\boldsymbol{\Sigma}}_{\mathbf p_j})$. Consequently, the measurement model of the active observation measurement can be derived as:

\begin{small}
    \begin{equation}\label{active_observation_model}
        {^{b_i}}\breve{\mathbf p}_{b_j} =  
        \underbrace{ \textcolor{black}{ \left( {^{G_i}}\mathbf T_{b_i}^{-1}  {^{G_i}}\mathbf T_{G_j} \right) \! \circ \!({^{G_j}}\breve{\mathbf p}_{b_j}  \! + \! \mathbf n_{\mathbf p_j} )}  \! + \! \mathbf n_{ao,ij} }_{\mathbf h_{ao,ij}(\mathbf x_{i},\mathbf n_{\mathbf p_j},\mathbf n_{ao,ij})} 
    \end{equation}
\end{small}
which defines a valid measurement equation about the state vector $\mathbf x_i$ containing ego-pose ${^{G_i}}\mathbf T_{b_i}$ and global extrinsic ${^{G_i}}\mathbf T_{G_j}$. The received teammate position ${^{G_j}}\breve{\mathbf p}_{b_j}$ is known, and $\mathbf n_{\mathbf p_j},\mathbf n_{ao,ij}$ are measurement noises.

Similarly, the explicit measurement model of the passive observation measurement $^{b_j}\breve{\mathbf p}_{b_i}$ for UAV $i$ can be obtained by projecting UAV $i$'s ground-true position $^{G_i}\mathbf p_{b_i}$ into UAV $j$'s body frame using the ground-true global extrinsic ${^{G_i}}\mathbf T_{G_j}$ and the ground-true ego-pose ${^{G_j}}\mathbf T_{b_j} = ({^{G_j}}\mathbf R_{b_j}, {^{G_j}}\mathbf p_{b_j})$ of UAV $j$. Then considering the measurement noise $\mathbf n_{po,ij} \sim \mathcal N(\mathbf 0,\boldsymbol{\Sigma}_{po,ij})$ of the passive observation measurement, the measurement model is:

\begin{small}
    \begin{equation}
        {^{b_j}}\breve{\mathbf p}_{b_i} = 
        \textcolor{black}{ \left( {^{G_j}}\mathbf T_{b_j}^{-1}  {^{G_i}}\mathbf T_{G_j}^{-1} \right) \! \circ \!{^{G_i}}\mathbf p_{b_i} }\! + \! \mathbf n_{po,ij}.
    \end{equation}
\end{small}

Since UAV $j$'s ground-true pose ${^{G_j}}{\mathbf T}_{b_j}$ is not a part of the state vector $\mathbf x_i$ as defined in \eqref{kinematic_model}, we similarly utilize the estimated ego-pose ${^{G_j}}\breve{\mathbf T}_{b_j}$ of UAV $j$ and the covariance $\breve{\boldsymbol{\Sigma}}_{\mathbf T_j}$ received from UAV $j$, to model the ground-true pose of UAV $j$ as $ ^{G_j}\mathbf T_{b_j} = {^{G_j}}\breve{\mathbf T}_{b_j} \textcolor{black}{\boxplus} \mathbf n_{\mathbf T_j}$, where the noise $\mathbf n_{\mathbf T_j} \sim \mathcal N(\mathbf 0, \breve{\boldsymbol{\Sigma}}_{\mathbf T_j})$.
Consequently, the passive observation measurement model is:

\begin{small}
    \begin{equation}\label{passive_observation_model}
        {^{b_j}}\breve{\mathbf p}_{b_i} = 
        \underbrace{ \textcolor{black}{ \left( ({^{G_j}}\breve{\mathbf T}_{b_j} \textcolor{black}{\boxplus} \mathbf n_{\mathbf T_j})^{-1}  {^{G_i}}\mathbf T_{G_j}^{-1} \right) \! \circ \! {^{G_i}}\mathbf p_{b_i}} \! + \! \mathbf n_{po,ij}}_{
        \mathbf h_{po,ij}(\mathbf x_{i},\mathbf n_{\mathbf T_j}, \mathbf n_{po,ij}),
    }
\end{equation}
\end{small}
which defines a valid measurement equation about the state vector $\mathbf x_i$ containing ego-position ${^{G_i}}\mathbf p_{b_i}$ and global extrinsic ${^{G_i}}\mathbf T_{G_j}$. The received teammate pose ${^{G_j}}\breve{\mathbf T}_{b_j}$ is known, and $\mathbf n_{\mathbf T_j},\mathbf n_{po,ij}$ are measurement noises.

To sum up, the entire measurement vector $\mathbf y$, the observation function $\mathbf h$ and the observation noise $\mathbf v$ (the subtract $k$ is omitted for simplification) are

\begin{small}
    \begin{equation}\label{obs_func_sum}
    \begin{aligned}
        \mathbf y &= \begin{bmatrix}
            \underbrace{\cdots,\mathbf{0}, \cdots}_{\text{point measurements}} , \underbrace{\cdots, {^{b_i}}\breve{\mathbf p}_{b_j}^T ,\cdots}_{\text{active observation measurements}},\underbrace{\cdots, {^{b_j}}\breve{\mathbf p}_{b_i}^T ,\cdots}_{\text{passive observation measurements}}
        \end{bmatrix}^T,\\
        \mathbf h &= \begin{bmatrix}
            \cdots,\mathbf h_{p,n}^T , \cdots,\cdots, \mathbf h_{ao,ij}^T ,\cdots,\cdots, \mathbf h_{po,ij}^T ,\cdots
        \end{bmatrix}^T, \\
        \mathbf v &= \begin{bmatrix}
            \cdots,\mathbf n_{p, n}^T  , \cdots ,\cdots, \mathbf n_{\mathbf p_j}^T , \mathbf n_{ao,ij}^T , \cdots ,\cdots, \mathbf n_{\mathbf T_j}^T  , \mathbf n_{po,ij}^T ,\cdots
        \end{bmatrix}^T. \\
    \end{aligned}
\end{equation}
\end{small}

\subsubsection{Temporal Compensation of Mutual Observation Measurements} \label{section:compensation}
For the measurement models \eqref{active_observation_model} and \eqref{passive_observation_model} to be valid, the involved states and measurements should be at the same time.
However, due to the asynchronous nature of state estimation among different UAVs and the presence of transmission delays, the states and measurements from different UAVs are usually asynchronous. 
Therefore, it is necessary to compensate for the temporal mismatch between the received measurements or states, and the ego-state in the measurement models. \textcolor{black}{While the previous work \cite{zhu2023swarm} ignores this temporal mismatch, this paper carefully addresses this problem based on a constant velocity model.}

For the active observation measurement model \eqref{active_observation_model}, the measurement ${^{b_i}}\breve{\mathbf p}_{b_j}$ is a cluster of points, which are undistorted and projected to the scan end time $t_{i,k}$ (see Section \ref{section:detection}). The received UAV $j$'s position ${^{G_j}}\breve{\mathbf p}_{b_j}$, however, is estimated at timestamp $t_{j,k}$ in UAV $j$'s system time. To make a valid measurement model at time $t_{i,k}$, UAV $j$'s position ${^{G_j}}\breve{\mathbf p}_{b_j}$ should be temporally compensated from its time of estimation (\ie, $t_{j,k}$) to the time the measurement model is established (\ie, $t_{i,k}$), according to a constant velocity model from its estimated velocity ${^{G_j}}\breve{\mathbf v}_{b_j}$: 

\begin{small}
\begin{equation}\label{compensation_ao}
    {^{G_j}}\breve{\mathbf p}_{b_j}^{\text{comp}} = {^{G_j}}\breve{\mathbf p}_{b_j} \! + \! {^{G_j}}\breve{\mathbf v}_{b_j} (t_{i,k}-t_{j,k} + {^i}\tau_j),
\end{equation}    
\end{small}
which should be substituted into \eqref{active_observation_model} to supply the original measurement ${^{b_i}}\breve{\mathbf p}_{b_j}$. The resultant measurement model with temporal compensation is hence:

\begin{small}
    \begin{equation} 
        \begin{aligned}
             {^{b_i}}\breve{\mathbf p}_{b_j} = &
            \textcolor{black}{ \left({^{G_i}}\mathbf T_{b_i}^{-1} {^{G_i}}\mathbf T_{G_j} \right)} \circ (
            {^{G_j}}\breve{\mathbf p}_{b_j} \\
             & +  {^{G_j}}\breve{\mathbf v}_{b_j} 
            (   t_{i,k}  - t_{j,k}  +  {^i}\tau_j  )
             +  \mathbf n_{\mathbf p_j}  )   +  \mathbf n_{ao,ij},   
        \end{aligned} 
    \end{equation}
\end{small}
which is a measurement equation about the state $\mathbf x_i$ containing ego-pose ${^{G_i}}\mathbf T_{b_i}$ and global extrinsic ${^{G_i}}\mathbf T_{G_j}$.

For the passive observation measurements model \eqref{passive_observation_model}, the passive observation measurement ${^{b_j}}\breve{\mathbf p}_{b_i}$ is transmitted from UAV $j$ and is estimated at timestamp $t_{j,k}$ of UAV $j$'s system time. To establish a valid measurement model at the time indicated by $t_{j,k}$, all the states and other measurements in \eqref{passive_observation_model} should also be at $t_{j,k}$. The received UAV $j$'s state ${^{G_j}}\breve{\mathbf T}_{b_j}$ is already stamped with $t_{j,k}$, while the ego-position ${^{G_i}}\mathbf p_{b_i}$, which is the state at $t_{i,k}$, can be compensated using a constant velocity model as follows:

\begin{small}
\begin{equation}\label{compensation_po}
    ^{G_i}\mathbf p_{b_i}^{\text{comp}} = {^{G_i}}\mathbf p_{b_i} + {^{G_i}}\mathbf v_{b_i} (t_{j,k} -t_{i,k} - {^i}\tau_j),
\end{equation}
\end{small}
which should be substituted into \eqref{passive_observation_model} to supply the original state ${^{b_j}}{\mathbf p}_{b_i}$. The temporally compensated measurement model is hence

\begin{small}
    \begin{equation}
        \begin{aligned}
                {^{b_j}}\breve{\mathbf p}_{b_i} = &
        ({^{G_j}}\breve{\mathbf T}_{b_j} \textcolor{black}{\boxplus} \mathbf n_{\mathbf T_j})^{-1} {^{G_i}}\mathbf T_{G_j}^{-1} \textcolor{black}{ \circ } (
        {^{G_i}}\mathbf p_{b_i} \\
        & +  {^{G_i}}\mathbf v_{b_i} (t_{j,k}  -  t_{i,k}  -  {^i}\tau_j))
          +  \mathbf n_{po,ij},
        \end{aligned}
    \end{equation}
\end{small}
which is a measurement equation about the state $\mathbf x_i$ containing ego-position ${^{G_i}}\mathbf p_{b_i}$, velocity ${^{G_i}}\mathbf v_{b_i}$, and global extrinsic ${^{G_i}}\mathbf T_{G_j}$.

\subsubsection{State and Covariance Update}\label{section:update}
Based on the LiDAR point measurement model \eqref{eq:point_measurement}, mutual observation models \eqref{active_observation_model} and \eqref{passive_observation_model} with temporal compensation explained previously, we leverage an iterated Kalman filter (ESIKF) \cite{he2021kalman} to update the state repeatedly. 
This process will repeat until convergence, then the optimal state estimation and covariance are obtained. The detailed computation of Kalman gain and update steps can be referred to \cite{xu2021fast,xu2022fast}.
After the update, covariance re-initialization will be implemented following Section \ref{section:marginalization} for the next round of state estimation.

\subsection{Marginalization}\label{section:marginalization}
The dimension of the state defined in \eqref{kinematic_model} would increase linearly with the swarm size, leading to an almost cubic growth of computation complexity of the ESIKF. To \textcolor{black}{address the problem of} explosion of \textcolor{black}{state dimension} and computational complexity \textcolor{black}{in the previous work \cite{zhu2023swarm}}, we propose a novel marginalization method. 
In the flight of aerial swarm systems, due to the restricted detecting range and FoV of the LiDAR sensor, the UAVs are typically unable to observe all teammate UAVs at all times. The observed teammates (either active or passive) have their global extrinsic transformations ${^{G_i}}\mathbf T_{G_j}$ persistently excited, as shown in \eqref{active_observation_model} and \eqref{passive_observation_model}, while others do not. Therefore, we only need to update the global extrinsic of teammate UAVs which can observe the self-UAV (contributing a passive observation measurement) or are observed by the self-UAV (contributing an active observation measurement). This is achieved by a marginalization operation below.

For simplification, we omit the subtract $k$ and $i$, which represent the $k$-th estimation of UAV $i$.
After receiving the $k$-th LiDAR scan, we identify the mutual observations as detailed in Section \ref{section:meas_modeling}. Let $\mathcal A$ denote the set of teammate UAVs that are observed in the current scan and $\mathcal B$ the set of teammate UAVs that are not. Let $\mathbf x_1$ represent the sub-state consisting of the ego-state and global extrinsic w.r.t. teammates in the set $\mathcal A$, while $\mathbf x_2$ represents the complementary state consisting of global extrinsic w.r.t. teammates in the set $\mathcal B$. We get $\dim(\mathbf x_1) = 18 + 6K$ and $\dim(\mathbf x_2) = 6(N-1-K)$, where $K$ represent the number of teammates with mutual observation (\ie, $\dim(\mathcal{A}) = 6K$). Furthermore, in the current round of state estimation, assume $(\widehat {\mathbf x},\widehat {\mathbf P})$ as propagated state and covariance after a normal ESIKF prediction step (\ie, Section \ref{section:propagation}). Then, they can be partitioned as:

\begin{small}
    \begin{equation}\label{eq:distribution}
\begin{aligned}
    \mathbf x \sim
    \mathcal N(
        \widehat {\mathbf x},
        \widehat {\mathbf P}
    )
    =
    \mathcal N(
    \begin{bmatrix}
        \widehat {\mathbf x}_1\\
        \widehat {\mathbf x}_2
    \end{bmatrix},
    \begin{bmatrix}
        \widehat {\boldsymbol \Sigma}_{11} & \widehat {\boldsymbol \Sigma}_{12}\\
        \widehat {\boldsymbol \Sigma}_{21} & \widehat {\boldsymbol \Sigma}_{22}\\
    \end{bmatrix}
    ).
\end{aligned}
\end{equation}
\end{small}

Since $\mathbf x_2$ will not be updated due to the lack of persistent excitation, we marginalize it out from $\mathbf x$, leading to the prior distribution of the two sub-states:

\begin{small}
    \begin{equation}\label{eq:marginal}
        \mathbf x_1 \sim\mathcal N(\widehat {\mathbf x}_1, \widehat {\boldsymbol \Sigma}_{11}),  \quad
        \mathbf x_2 \sim \mathcal N(\widehat {\mathbf x}_2, \widehat {\boldsymbol \Sigma}_{22}).
    \end{equation}
\end{small}

To update the sub-state $\mathbf x_1$, we notice the measurement model

\begin{small}
    \begin{equation}\label{eq:meas_model_x1}
        \mathbf y = \mathbf h(\mathbf x,\mathbf v) = \mathbf h(\mathbf x_1,\mathbf v_1),
    \end{equation}
\end{small}
where $\mathbf y$ includes point measurements and mutual observation measurements (both active and passive), which depend only on $\mathbf x_1$. Then, $\mathbf x_1$ can be updated by fusing the prior distribution $\mathbf x_1 \sim \mathcal N(\widehat {\mathbf x}_1, \widehat {\boldsymbol \Sigma}_{11})$ with the measurements $\mathbf y$ by following the normal ESIKF update step (\ie, Section \ref{section:update}). Assume the updated state estimate and covariance are $\bar{\mathbf x}_1$ and $\bar{\boldsymbol{\Sigma}}_{11}$ respectively. Then, we have $\mathbf x_1 \sim \mathcal N(\bar {\mathbf x}_1, \bar {\boldsymbol \Sigma}_{11})$ and that the sub-state $\mathbf x_2$ still remains at $\mathbf x_2 \sim \mathcal{N}(\widehat{\mathbf x}_2, \widehat{\boldsymbol \Sigma}_{22})$. Now that $\mathbf x_1$ and $\mathbf x_2$ are two independent distributions, they should evolve separately in the subsequent ESIKF steps. Specifically, for $\mathbf x_2$,  it is subject to its state transition function:

\begin{small}
    \begin{equation}\label{eq:x_2}
        \mathbf x_{2,\tau +1} = \mathbf x_{2,\tau}
    \end{equation}
\end{small}
while for $\mathbf x_1$, it is subject to 

\begin{small}
    \begin{equation}\label{eq:x_1}
        \mathbf x_{1,\tau+1} = \mathbf x_{1,\tau} \boxplus(\Delta t_\tau \mathbf f_1(\mathbf x_{1,\tau}, \mathbf u_{\tau}, \mathbf w_{\tau})),
    \end{equation}
\end{small}
where $\mathbf f_1(\mathbf x_{1,\tau}, \mathbf u_{\tau}, \mathbf w_{\tau})$ takes the first $18 + 6K$ elements of $\mathbf f(\mathbf x_{\tau}, \mathbf u_{\tau}, \mathbf w_{\tau})$ in \eqref{transition_func}.  

In the next round of ESIKF, each of the two sub-state will propagate starting from their respective initial distribution, $\mathbf x_1 \sim \mathcal N(\bar {\mathbf x}_1, \bar {\boldsymbol \Sigma}_{11})$ and $\mathbf x_2 \sim \mathcal N(\widehat {\mathbf x}_2, \widehat {\boldsymbol \Sigma}_{22})$, and following their respective state transition function \eqref{eq:x_1} and \eqref{eq:x_2}. This process can be expressed compactly by propagating the complete system following \eqref{discrete_model} from an initial distribution $\mathbf x \sim \mathcal{N}(\bar{\mathbf x}, \bar{\mathbf P})$ defined below

\begin{small}
    \begin{equation}
        \bar{\mathbf x} = 
        \begin{bmatrix}
            \bar{\mathbf x}_1\\ 
            \widehat{\mathbf x}_2
        \end{bmatrix},
        \bar {\mathbf P} =  
        \begin{bmatrix}
        \bar{\boldsymbol\Sigma}_{11} & \mathbf 0\\
        \mathbf 0 & \widehat{\boldsymbol\Sigma}_{22}\\
        \end{bmatrix}.
    \end{equation}
\end{small}

Packing the posterior distribution $\mathbf x_1 \sim \mathcal N(\bar {\mathbf x}_1, \bar {\boldsymbol \Sigma}_{11})$ and the prior distribution $\mathbf x_2 \sim \mathcal N(\widehat {\mathbf x}_2, \widehat {\boldsymbol \Sigma}_{22})$ into the joint distribution $\mathbf x \sim \mathcal{N}(\bar{\mathbf x}, \bar{\mathbf P})$ is termed as ``covariance re-initialization". With the covariance re-initialization, the propagation of the next step can simply follow the standard ESIKF prediction step of the complete system \eqref{discrete_model}, which is detailed in Section \ref{section:propagation}.

\subsection{Degeneration Evaluation}
\label{section:degeneration_detection}
The ESIKF presented previously would update the global extrinsic of teammate UAVs along with the ego-state. However, the update is valid only when the LiDAR scan contains sufficient geometric features. In some extreme environments, LiDAR sensors may encounter degeneration where the point-cloud fails to provide sufficient constraints to determine its ego-pose, making it impossible to distinguish the global extrinsic from ego-motion given mutual observation measurements, \textcolor{black}{which is a problem suffered by our previous work \cite{zhu2023swarm}. To address this problem, we propose to automatically detect LiDAR degeneration. If it occurs}, the previously estimated global extrinsic is used with mutual observation measurements to provide constraints for determining the ego-pose. The switching between the two cases (\ie, updating global extrinsic along with ego-state, and, using currently-estimated global extrinsic for ego-state update) can be achieved automatically by leveraging the marginalization operation as follows.

When LiDAR degeneration occurs, we marginalize all global extrinsic out from the state vector by setting $\mathcal A$ to null and $\mathcal B$ to the full set of all teammate UAVs. Thus, sub-state $\mathbf x_1$ only includes ego-state with $\dim(\mathbf x_1) = 18$, and $\mathbf x_2$ contains the global extrinsic transformation w.r.t. to all the teammates with $\dim(\mathbf x_2) = 6(N-1)$.
For the measurement model \eqref{eq:meas_model_x1}, we rewrite it as 

\begin{small}\color{black}
    \begin{equation}
        \mathbf y = \mathbf h(\mathbf x, \mathbf v) =\mathbf h(\mathbf x_1, \mathbf x_2, \mathbf v) = \mathbf h(\mathbf x_1, \underbrace{[\mathbf x_2, \mathbf v]}_{\mathbf v_{\text{ext}}}) = \mathbf h(\mathbf x_1, \mathbf v_{\text{ext}}),
    \end{equation}
\end{small}
where \textcolor{black}{the marginalized sub-state $\mathbf x_2$ is an exogenous random signal (\ie, it is independent of the state $\mathbf x_1$) just like the measurement noise $\mathbf v$, so it is grouped with $\mathbf v$ to form the extended measurement noise $\mathbf v_{\text{ext}}$.} The distribution of the ``measurement noise" $\mathbf x_2$ is \textcolor{black}{obtained by propagating its sub-system following \eqref{eq:x_2}.} The rest of the steps, including the update of $\mathbf x_1$ and the subsequent prediction step, will be identical to that in Section \ref{section:marginalization}. 
Besides the marginalization above, the mutual observation noise $\mathbf v = [\mathbf n_{ao,ij}, \mathbf n_{po,ij}]$ (see Section \ref{section:meas_modeling}) of UAV $i$ would be adjusted to a smaller value to provide adequate constraints for ego-pose determination.

To achieve the aforementioned operations, a degeneration evaluation module is required. Inspired by \cite{zhen2017robust,zhen2019estimating}, we evaluate the degeneration situation of UAV $i$ by implementing singular value decomposition (SVD) of the Jacobian matrix $\mathbf J_{\mathbf T}$ of $\mathbf h_{p,n}(\mathbf x_i, \mathbf 0)$ in \eqref{eq:point_measurement} w.r.t. the ego-pose ${^{G_i}}\mathbf T_{b_i}$:

\begin{small}
    \begin{equation}
             \mathbf J_{\mathbf T} = 
             \begin{bmatrix}
                 - \mathbf u_n^T {^{G_i}}\mathbf R_{b_i} \lfloor {^{b_i}}\mathbf p_n \rfloor_\wedge & \mathbf u_n^T
             \end{bmatrix},
    \end{equation}
\end{small}
where the notation $\lfloor \mathbf a \rfloor_\wedge$ represents the skew-symmetric matrix of vector $\mathbf a \in \mathbb R^{3\times1}$
that maps the cross-product operation.
By calculating the singular values of $\mathbf J_{\mathbf T}$, finding the smallest one $\lambda$, and comparing it with a predefined degeneration threshold $\epsilon_d$, we can obtain the evaluation result. If $\lambda < \epsilon_d$, UAV $i$ is regarded as encountering LiDAR degeneration, and the corresponding responses mentioned above will be activated for this round of updates.

It is worth mentioning that the proposed method can achieve automatic switchover between the two modes: when degeneration is detected, the global extrinsic transformations and mutual observation measurements are utilized to accurately determine the ego-state; when no degeneration occurs, the point-cloud measurements of LiDAR are used to refine the global extrinsic states.

\subsection{\textcolor{black}{Broadcast of State Estimation Results}}
\label{section:broadcast_of_state}
\textcolor{black}{After state estimation completes, the results including updated ego-pose $^{G_i} \bar{\mathbf T}_{b_i}$, velocity $^{G_i} \bar{\mathbf v}_{b_i}$, pose covariance $\bar{\mathbf P}_i$, and refined global extrinsic transformations $^{G_i} \bar{\mathbf T}_{G_j}$ are shared with all teammate UAVs through the decentralized Ad-Hoc network. The ego-pose and velocity sent to teammates are utilized for their mutual state estimation following \eqref{eq:mutual-state-est}. The ego-pose, pose covariance, and refined extrinsic transformations are sent to teammates to construct their mutual observation measurements (Section \ref{section:meas_modeling}) for the next step estimation. }

\textcolor{black}{\textbf{Remark 3:} The broadcast of the estimation results will also cause the refined global extrinsic transformations to be shared with a new UAV joining the swarm in the middle of a mission. The shared extrinsic will trigger the factor graphs of the new UAV to be updated, by inserting the refined extrinsic transformations received from the network. Optimizing the factor graph will then obtain the extrinsic between the new UAV and existing swarms. On the other hand, the shared extrinsic transformations will not trigger any factor graph update of existing UAVs in the swarm, as this edge has already existed in the factor graph.}

\section{Simulation Evaluation}

In this section, we conducted simulation experiments to evaluate the performance of the Swarm-LIO2 framework.

\subsection{Simulator Setup}
\begin{figure}[htbp]
	\setlength\abovecaptionskip{-0.05\baselineskip}
	\centering
	\includegraphics[width=0.9\linewidth]{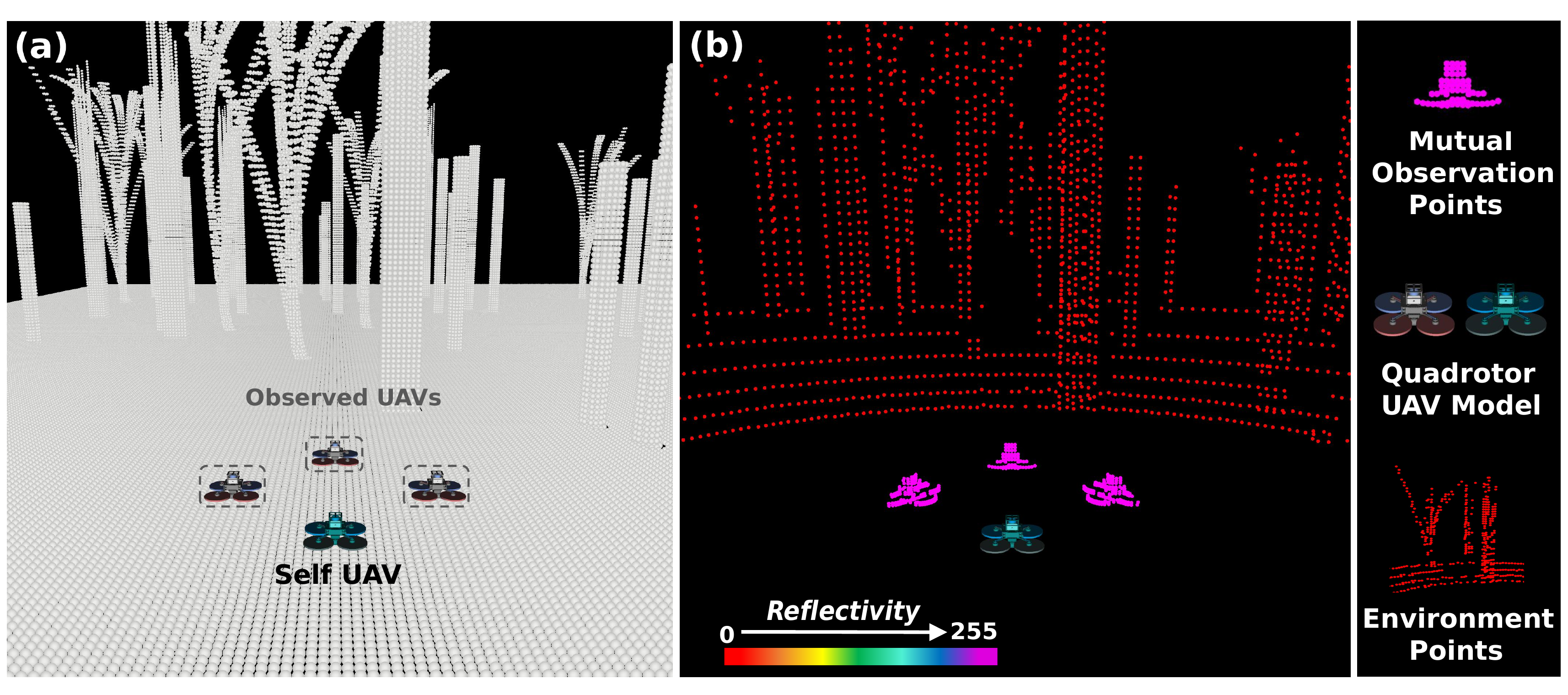}
	\caption{Illustration of MARSIM simulator and the corresponding rendered point-cloud. (a) Multi-UAV simulator scenario. (b) The rendered point-cloud of self-UAV colored by each point's reflectivity.} 
	\label{fig:sim} 
\end{figure}

In our simulation experiments, we utilize the MARSIM simulator\cite{kong2023marsim}, a lightweight point-realistic simulator for LiDAR-based UAVs. As shown in Fig.~\ref{fig:sim}, MARSIM supports a variety of common LiDAR models and we select Livox LiDAR sensors including Livox Avia and Livox Mid360 to maintain consistency with real-world experiment setup. It is worth mentioning that MARSIM is capable of simulating the mutual observation scenarios among multiple UAVs, which is essential for validating the method proposed in this paper.
To simulate the scenario where each UAV, in reality, is equipped with reflection tapes, the reflectivity of the mutual observation points observed by each UAV is set to large saturated values. In all simulation experiments, the simulator is running on a laptop with i9-12900H CPU and NVIDIA GeForce RTX 3080 Ti GPU, and the LiDAR scan rate is set to \SI{10}{Hz}. 
\vspace{-3mm}

\subsection{Initialization Efficiency \textcolor{black}{and Accuracy} Evaluation}
A key step in the initialization of an aerial swarm is the global extrinsic calibration. In our previous work, Swarm-LIO \cite{zhu2023swarm}, the identification and global extrinsic calibration are achieved solely through trajectory matching, necessitating each UAV to fly a certain trajectory. Each UAV performing this initialization trajectory in turn will lead to successive long flight distances, especially when the swarm size is large. 
In contrast, the decentralized pose graph optimization in Swarm-LIO2 requires only one UAV to fly a certain trajectory that can be observed by teammate UAVs, significantly reducing the initialization complexity. 

We validate the initialization efficiency by comparing it to Swarm-LIO \cite{zhu2023swarm} at swarm size varying from 0 to 40. At each swarm size, for Swarm-LIO2, only one UAV executes a figure-8 trajectory that can be observed by the rest UAVs. For Swarm-LIO, each drone needs to fly \textcolor{black}{the figure-8} trajectory in other UAVs' FoV.
Table~\ref{table:distance} shows the total flight distance for all UAVs in the initialization at different swarm sizes. As can be seen, as the swarm size increases, the total flight distance in \cite{zhu2023swarm} increases linearly, while that of Swarm-LIO2 increases very slowly and nearly remains unchanged regardless of the number of UAVs.
This indicates that the proposed method effectively mitigates the need for individual UAVs to fly extensive distances during initialization, compared to \cite{zhu2023swarm}. This contributes to significant energy savings and increased effective operational flight time for the swarm system.

\textcolor{black}{We also evaluate the initialization accuracy of Swarm-LIO2 and Swarm-LIO \cite{zhu2023swarm} using their respective initialization trajectories (\ie, only one UAV flies a figure-8 trajectory for Swarm-LIO2 versus all UAVs fly figure-8 trajectories for Swarm-LIO). 
By comparing the RMSE of the global extrinsic transformations obtained by Swarm-LIO2 with Swarm-LIO\cite{zhu2023swarm} at swarm size varying from 0 to 40, it can be observed from Table~\ref{table:init_accuracy} that the two methods have similar initialization accuracy, which means the proposed factor graph optimization nearly does not deteriorate the initialization accuracy.}

\begin{table}[tbp]
    \centering
    \renewcommand\arraystretch{1.3}
    \caption{Total Flight Distance in Initialization}
    \label{table:distance}
    \scalebox{0.85}{
    \begin{tabular}{c c c c c c c}
    \toprule
        \diagbox{Method}{Distance (m)}{Size} &5 &10 & 15 &20 &30 & 40\\ 
        \hline
        \textbf{Swarm-LIO2} &\textbf{23.2} &\textbf{23.2} &\textbf{23.2} &\textbf{23.2} &\textbf{30.7} & \textbf{43.5}\\
        \text{Swarm-LIO} &120.5 & 243.8 & 367.1 & 496.6 & 763.1 & 1032.2 \\
        \hline
    \end{tabular}
    }
     \vspace{-2mm}
\end{table}

\begin{table}[tbp]\color{black}
\centering
    \renewcommand\arraystretch{1.3}
    \caption{Initialization Accuracy Comparison}
    \label{table:init_accuracy}
    \scalebox{0.85}{
    \begin{tabular}{p{0.6cm}<{\centering}  p{1.8cm}<{\centering}   p{0.7cm}<{\centering}   p{0.7cm}<{\centering}   p{0.7cm}<{\centering} p{0.7cm}<{\centering}   p{0.7cm}<{\centering}  p{0.7cm}<{\centering}}
    \toprule
         RMSE & \diagbox{Algo.}{Size} & 5 & 10 & 15 & 20 & 30 & 40\\ 
          \cline{1-8}
        \multirow{2}* {\makecell[c]{\textbf{Trans} \\ \textbf{(m)}} } &
        \textbf{Swarm-LIO2}  & \textbf{0.1035}  &  0.1206   &  \textbf{0.1138}   &  \textbf{0.1395}    & 0.1323    & 0.1547   \\
                 & Swarm-LIO & 0.1088  &   \textbf{0.1193}   &  0.1195    & 0.1440   &    \textbf{0.1264}   & \textbf{0.1496}  \\ 
        \multirow{2}* {\makecell[c]{\textbf{Rot} \\ \textbf{(rad)}} } &
        \textbf{Swarm-LIO2} & \textbf{0.0623}  & 0.0739  &  0.0684 & \textbf{0.0717}  & 0.0860  &  0.0848 \\
                & Swarm-LIO & 0.0652 &   \textbf{0.0698}  &  \textbf{0.0644}   &   0.0762    & \textbf{0.0813}  &  \textbf{0.0821}\\
        \bottomrule
    \end{tabular}
    }
     \vspace{-2mm}
\end{table}

\subsection{State Estimation and Global Extrinsic Accuracy Evaluation}

\begin{figure}
        \setlength\abovecaptionskip{-0.05\baselineskip}
	\centering
	\begin{minipage}[t]{\linewidth}
		\centering
		\includegraphics[width=0.9\linewidth]{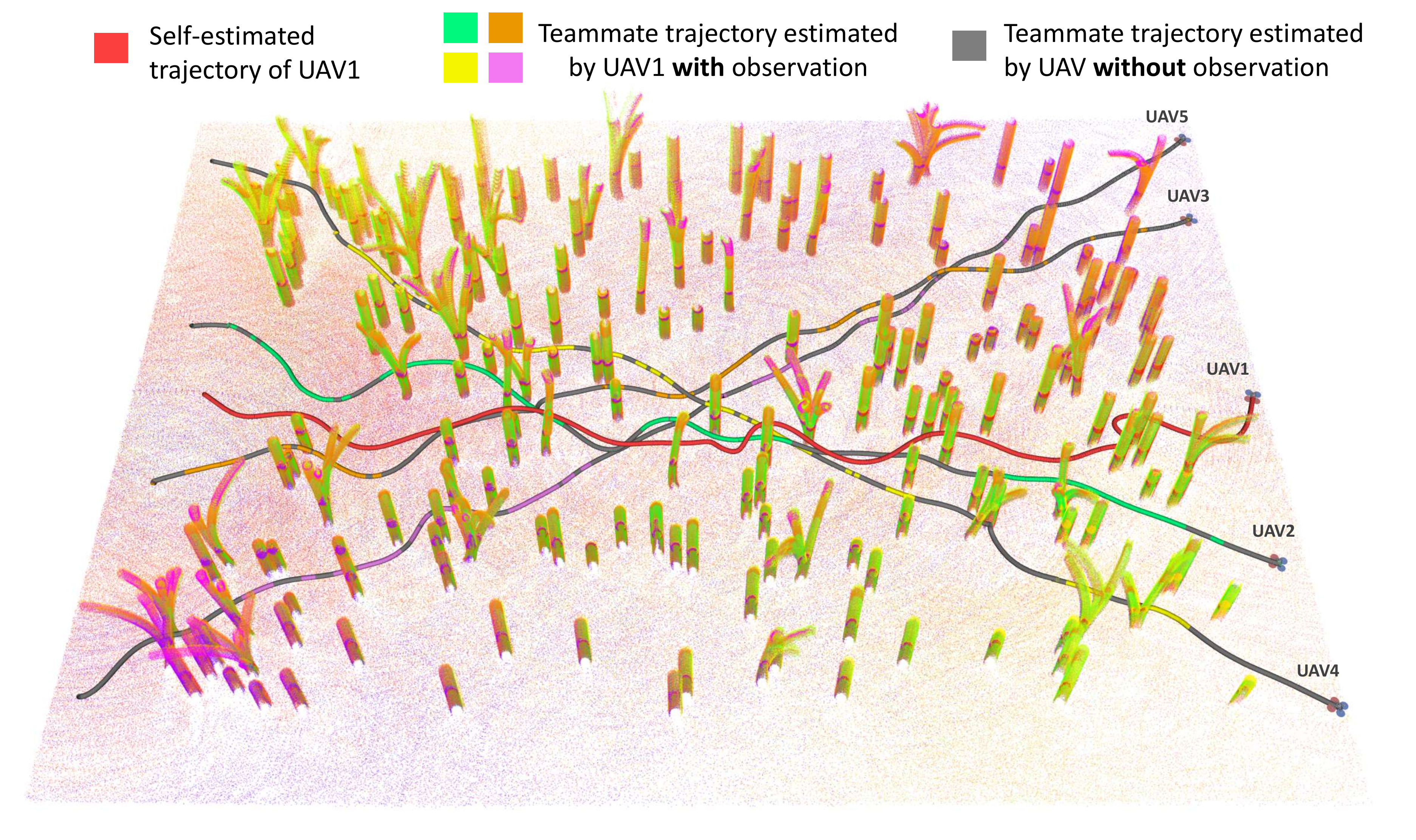}
		\caption{The estimated ego \textcolor{black}{trajectory} and  \textcolor{black}{teammate trajectories} on UAV1 in a five UAVs swarm system. The point-cloud map is generated in a post-processing stage where the maps of different UAVs are merged using the estimated global extrinsic transformations.}
		\label{fig:mutual_estimate}
        \vspace{3mm} 
	\end{minipage}
	\\
	\begin{minipage}[t]{\linewidth}
		\centering
		\includegraphics[width=0.95\linewidth]{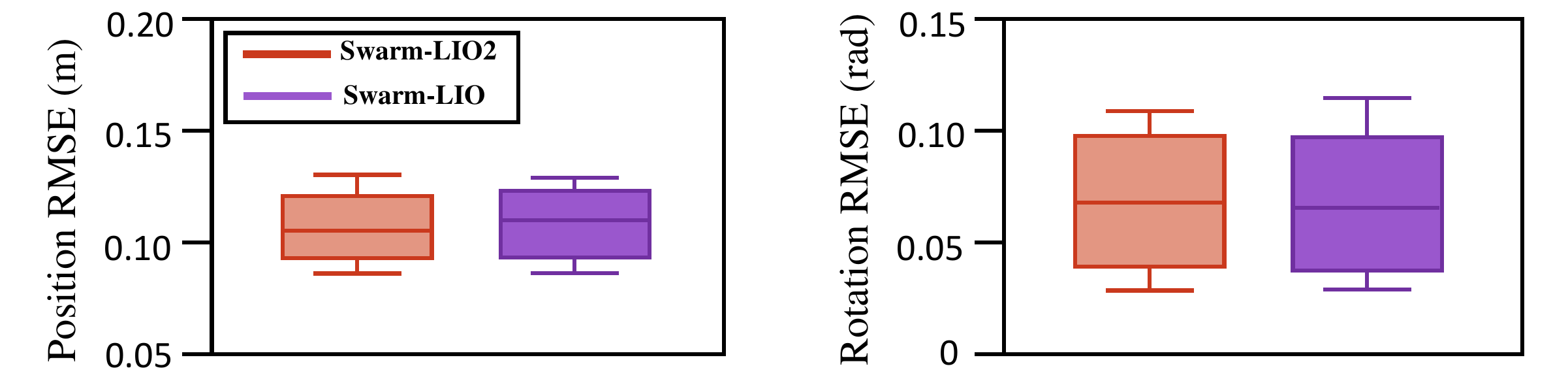}
		\caption{The error (RMSE) distribution of the five UAVs' trajectories estimated by different methods. }
		\label{fig:error_distribution}
        \vspace{3mm} 
	\end{minipage}
        \\
	\begin{minipage}[t]{\linewidth}
		\centering
		\includegraphics[width=0.95\linewidth]{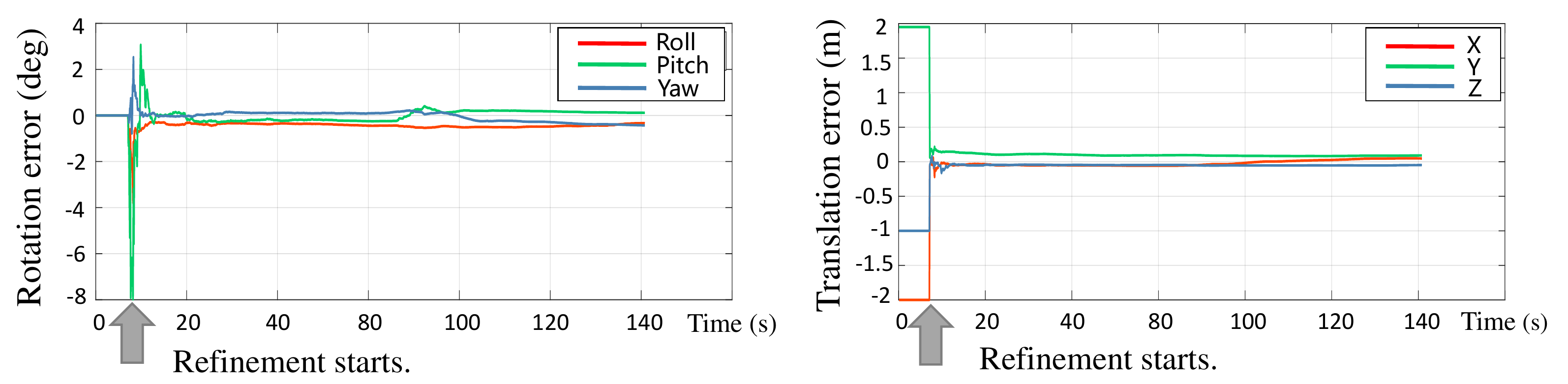}
		\caption{The global extrinsic estimation error of UAV1 w.r.t. UAV2 (\ie, ${^{G_1}}\mathbf R_{G_2}, {^{G_1}}\mathbf p_{G_2}$).}
		\label{fig:global_extrinsic}
	\end{minipage}
 \vspace{-2mm}
\end{figure}

Swarm-LIO2 can achieve robust, accurate ego-state and mutual state estimation and provide effective global extrinsic transformation.
This capability is indispensable for various swarm applications, such as multi-UAV formation flight, mutual collision avoidance, collaborative exploration, etc. As explained in Section \ref{section:mutual_state_estimation}, the mutual state estimation is robust to mutual observation loss. To evaluate such performance, the simulation experiments are conducted in a randomly generated 3D forest-like scenario of dimension $60 \times 40 \times 8~\mathrm{m}^3$ with a swarm composed of 5 UAVs (Fig.~\ref{fig:mutual_estimate}). In this evaluation, each UAV needs to perform ego-state estimation as well as mutual state estimation of the other four UAVs in the simulated forest. After the swarm initialization, the UAVs fly through the forest from one side to the other, causing frequent mutual observation losses between any two UAVs due to the dense obstacles. As shown in Fig.~\ref{fig:mutual_estimate} in which UAV1 is selected as the self-UAV, despite the frequent mutual observation losses caused by occlusions, the ego  \textcolor{black}{trajectory} and  \textcolor{black}{teammate trajectories} estimated by Swarm-LIO2 on UAV1 can maintain smoothness and continuity. \textcolor{black}{We also transform the point cloud maps constructed by each UAV to the global frame of UAV1, using the estimated global extrinsic transformations. As can be seen, the merged point cloud map maintains a high level of consistency, which qualitatively showcases the excellent accuracy of the global extrinsic estimation.}

For quantitative evaluation, we compute the error (RMSE) of all the estimated UAV trajectories by comparing them to the ground-truth offered by the simulator. Since each of the $N$ UAV trajectories is estimated $N$ times by itself and the rest of the teammates, we compute the RMSE of all the $N^2$ estimated trajectories and compare them with Swarm-LIO\cite{zhu2023swarm}. The distribution of all the $N^2$ RMSE, separated by position and rotation, are illustrated in Fig.~\ref{fig:error_distribution}. It can be observed that compared to Swarm-LIO, Swarm-LIO2 achieves a similar accuracy despite the introduced marginalization operations.

Finally for the global extrinsic estimation, Fig.~\ref{fig:global_extrinsic} shows the initialization and online refinement of the global extrinsic transformation, we select UAV1 and UAV2 for analysis and depict the error of the estimated global extrinsic ${^{G_1}}\mathbf R_{G_2}, {^{G_1}}\mathbf p_{G_2}$, in which the ground-truth is provided by the simulator. 
It can be seen that the estimation error gradually converges during the online refinement, and the final error of the global extrinsic is less than 1$^\circ$ (for rotation) and \SI{0.2}{m} (for translation).
With the accurate global extrinsic, we can merge the point-cloud map produced by different UAVs, which is extremely useful for large-scale collaborative mapping. As shown in Fig.~\ref{fig:mutual_estimate}, all the point-cloud maps (points of teammate UAVs are filtered as they are dynamic) are transformed into UAV1's global frame using the estimated global extrinsic transformations. 
The consistently aligned map shown in Fig.~\ref{fig:mutual_estimate} indicates the accurate global extrinsic estimation of Swarm-LIO2.

\subsection{\textcolor{black}{Scalability Analysis in Time Consumption and Communication Bandwidth}}\label{section:time_consumption}

\begin{figure*}[tbp]
	\setlength\abovecaptionskip{-0.05\baselineskip}
	\centering
	\includegraphics[width=0.99\linewidth]{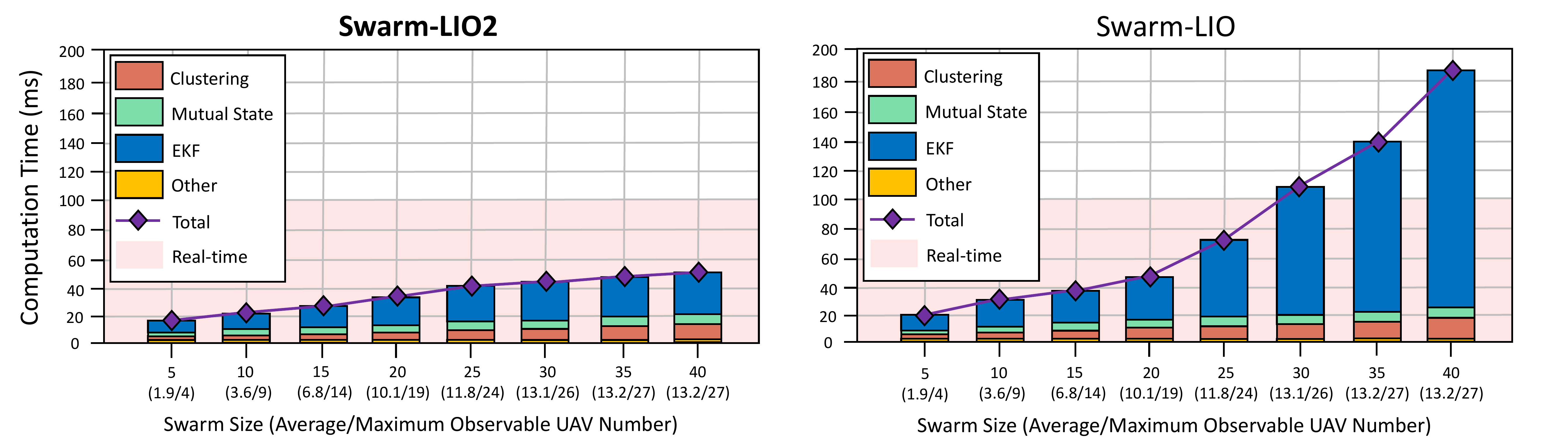}
	\caption{The overall and module-specific computation times per LiDAR scan under swarm scales varying from 0 to 40. \textcolor{black}{The x-label shows the swarm size along with the average/maximum teammates observed by any UAV in the swarm, which are shown in the parenthesis.}}
	\label{fig:time_consumption} 
\end{figure*}

To validate that Swarm-LIO2 possesses high scalability and can maintain efficient computation even at a large swarm size, a comparative experiment is conducted in a sparse simulated 3D forest-like scenario.
We compare the time consumption of Swarm-LIO2 to Swarm-LIO\cite{zhu2023swarm} at different swarm sizes. The entire framework of each method can be partitioned into several modules, including point clustering, mutual state estimation, ESIKF-based state estimation, etc. We analyze the time consumption of each module of the two methods and the results are shown in Fig.~\ref{fig:time_consumption}.

As can be seen, the computation time of clustering and mutual state estimation in the two methods linearly increases as the swarm size increases. This is because these two sub-modules need to be performed for every teammate UAV in the swarm system.
Moreover, since Swarm-LIO2 employs Fast Euclidean Clustering (FEC)\cite{cao2022fec} for clustering, which is extremely faster compared to traditional Euclidean clustering provided by the PCL library used in Swarm-LIO, the overall time consumption of clustering in Swarm-LIO2 is lower than that in Swarm-LIO.
For the ESIKF-based state estimation module, its time complexity is cubic to the state dimension in theory. In Swarm-LIO \cite{zhu2023swarm}, the state contains the ego-state and the global extrinsic transformations of all teammates, leading to a time consumption rapidly increasing with the swarm size. By contrast, in Swarm-LIO2, the state only includes the ego-state as well as the global extrinsic of observed teammates or teammates observing the self-UAV, which \textcolor{black}{often saturates at a relatively small number due to mutual occlusions and LiDAR FoV limit (see Fig.~\ref{fig:time_consumption})}. 
As a result, as the swarm size increases, the time consumption of Swarm-LIO2 increases sub-linearly and at a rather low rate, even exhibiting a saturation trend when the swarm size reaches a certain size. 
To sum up, Swarm-LIO2 is highly scalable compared to Swarm-LIO in terms of time consumption, reducing \SI{7.83}{ms}, \SI{31.65}{ms}, \SI{133.09}{ms} total consumed time at swarm size of 10, 25, and 40, respectively.


\begin{figure}[tbp]
	\setlength\abovecaptionskip{-0.05\baselineskip}
	\centering
	\includegraphics[width=0.95\linewidth]{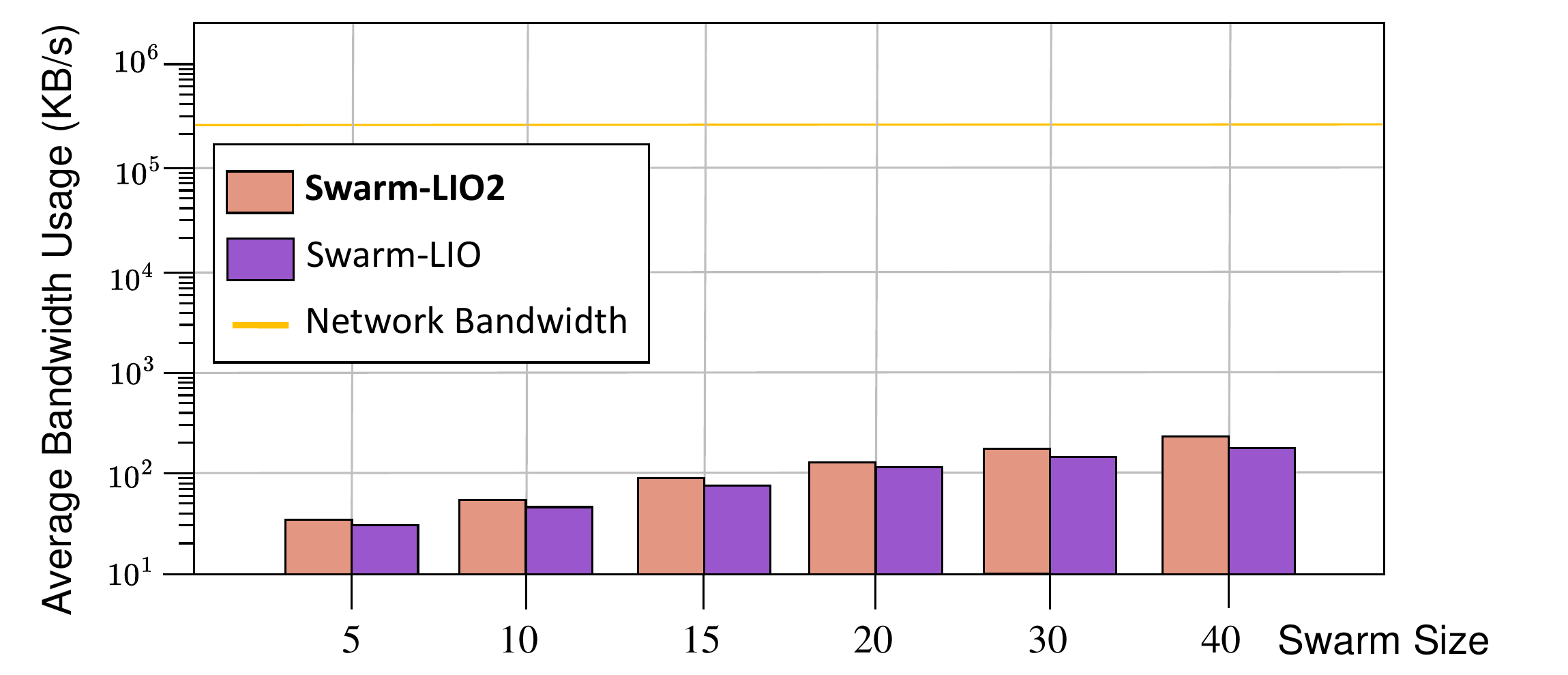}
	\caption{The average transmitting bandwidth usages (KB/s) of Swarm-LIO2 and Swarm-LIO at different swarm sizes.} 
	\label{fig:bandwidth} 
\vspace{-3mm}
\end{figure}

\textcolor{black}{In addition to computational time consumption, the communication overhead might be another bottleneck that prevents the swarm scale from growing.
Therefore, it is crucial to evaluate the communication bandwidth usage under different swarm scales. We count the average data transfer volume per second, which is the average transmitting bandwidth usage, of Swarm-LIO2 and the previous version Swarm-LIO\cite{zhu2023swarm}, at swarm size varying from 0 to 40. From the results shown in Fig.~\ref{fig:bandwidth}, it can be observed that the average bandwidth usages of both methods increase linearly as the swarm size grows, but still remains at a low level since all the information to be communicated is of low dimension. When the swarm size is 40, the bandwidth usage of Swarm-LIO2 is \textcolor{black}{below \SI{250}{KB/s}}. Compared to the bandwidth of the Intel Wi-Fi 6E AX211 (Gig+) \footnote{https://www.intel.cn/content/www/cn/zh/products/sku/204837/intel-wifi-6e-ax211-gig/specifications.html} adapter used in our real systems, which is \SI{2.4}{Gbps} (approximately \SI{300}{MB/s}), the bandwidth usage of Swarm-LIO2 is almost negligible, indicating that the bandwidth is not a bottleneck at all. Besides, the transmitting bandwidth usage of Swarm-LIO2 is slightly larger than that of Swarm-LIO because there is additional information (\eg, global extrinsic transformation) to be exchanged in Swarm-LIO2.}

\subsection{Fly Through a Degenerated Corridor}
In this section, we conduct a simulation experiment in which five UAVs equipped with Livox Mid360 LiDARs need to fly through a degenerated corridor. In this case, the measurements of a single LiDAR can not provide sufficient constraints for pose determination, but Swarm-LIO2 can perform robust and stable state estimation thanks to the mutual observation measurements from teammates. \textcolor{black}{We compare the localization accuracy of our method to Swarm-LIO and some state-of-the-art LiDAR-inertial odometry for a single UAV system and the results showcase the superior robustness of Swarm-LIO2 to degenerated scenes.
Due to the page limit, we put the detailed descriptions, illustrations, qualitative and quantitative results in the Supplementary
Material \cite{zhu2024supplementary}.}

\color{black}
\subsection{Localization Accuracy with Communication Loss}
The wireless communication is assumed to be perfect in all the previous simulation tests. 
However, in reality, communication issues like dropouts are inevitable since various interference sources, \eg, electromagnetic interference and physical occlusions would impact communication stability. In case of communication loss, Swarm-LIO2 would still hold the connection status for two more seconds (Section \ref{section:sync}), during which the teammates' states are predicted via (\ref{eq:mutual-state-est}) using the constant velocity model and the last updated extrinsic. Once the teammate connection status changes to “disconnected”, the teammate states will no longer be estimated until the teammate is reconnected. Holding the connection status for two more seconds can effectively reduce the false alarm caused by temporary communication loss such as temporary network congestion. Regardless of the communication loss, Swarm-LIO2 can reliably estimate the ego-state based on the measurements of LiDAR and IMU. In the case of complete communication loss, Swarm-LIO2 would degrade to FAST-LIO2\cite{xu2022fast}, to estimate the ego-state only.

To validate the robustness of Swarm-LIO2 to communication loss, we evaluate the state estimation accuracy on a swarm composed of five UAVs in the simulation environment shown in Fig.~\ref{fig:mutual_estimate}, under different simulated packet loss rates (PLRs). We evaluate the accuracy by averaging the RMSEs of the $N^2$ trajectories, which are shown in Table \ref{table:accuracy_plr}. As can be seen, as PLR increases, the localization accuracy of Swarm-LIO2 does not deteriorate obviously, which clearly illustrates the remarkable robustness of Swarm-LIO2 to communication dropouts.

\begin{table}[tbp]\color{black}
\centering
    \renewcommand\arraystretch{1.3}
    \caption{State Estimation Accuracy under Different PLRs}
    \label{table:accuracy_plr}
    \scalebox{0.85}{
    \begin{tabular}{p{3.4cm}<{\centering}  p{0.8cm}<{\centering}   p{0.8cm}<{\centering}   p{0.8cm}<{\centering} p{0.8cm}<{\centering} p{0.8cm}<{\centering}  }
    \toprule
         \diagbox{Avg. RMSE}{PLR(\%)} & 0 & 25 & 50 & 75 & 100\\ 
          \cline{1-6}
        Position (m) & 0.0754 & 0.0772 & 0.0851& 0.0882& 0.0865 \\
        Rotation (rad) & 0.0446& 0.0489& 0.0515& 0.0526& 0.0523\\
        \bottomrule
    \end{tabular}
    }
    \begin{tablenotes}
        \footnotesize
        \item
        Note: the average RMSE is not calculated for the time in which the corresponding teammate is marked as ``disconnected", during which the teammate state is not estimated. 
     \end{tablenotes}
     \vspace{-4mm}
\end{table}

\color{black}
\section{Real-world Applications}\label{section:real_exp}
To comprehensively demonstrate the properties of the proposed swarm state estimation method and its capability to support different applications, we conduct various experiments in real-world environments.

\subsection{Hardware Platform}
\begin{figure}[tbp]
	\setlength\abovecaptionskip{-0.05\baselineskip}
	\centering
	\includegraphics[width=0.95\linewidth]{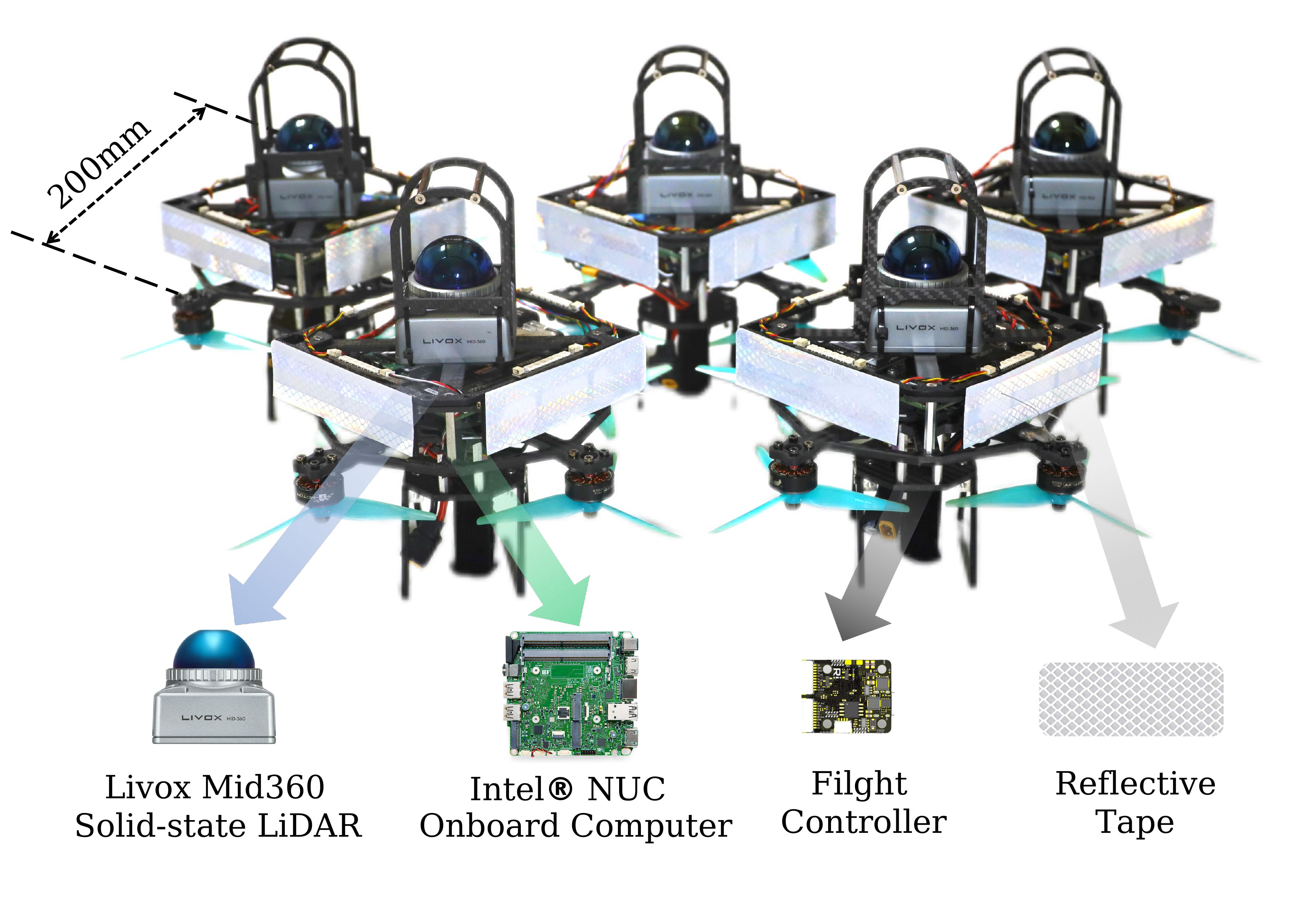}
	\caption{The UAV platform of the proposed swarm system. Each UAV is equipped with a 3D LiDAR, a flight controller with built-in IMU, and an onboard Intel NUC computer. Several reflective tapes are attached to the UAVs for teammate detection.} 
	\label{fig:platform} 
\vspace{-4mm}
\end{figure}

The experiment platform is a compact and cost-effective quadrotor UAV that is equipped with 3D LiDAR and IMU sensors.
The quadrotor UAV has a \SI{280}{mm} wheelbase and is equipped with a Livox Mid360 LiDAR. The LiDAR is capable of generating point clouds at a rate of 200,000 points per second and possesses 360$^\circ$ $\times$ 59$^\circ$ field of view. As for the computation unit, each UAV is equipped with an onboard Intel NUC computer featuring an i7-1260P CPU, coupled with a flight controller that provides over \SI{200}{Hz} IMU measurements. In all the real-world experiments, the LiDAR scan rate is \SI{30}{Hz}. Each UAV is attached with reflective tapes for easy detection.
The spatio-temporal extrinsic of the LiDAR and IMU are pre-calibrated with \cite{zhu2022robust}.
The hardware platform of our swarm system is shown in Fig.~\ref{fig:platform}.

\subsection{Inter-UAV Collision Avoidance}\label{section:inter-UAV}
\begin{figure*}[tbp]
	\setlength\abovecaptionskip{-0.05\baselineskip}
	\centering
	\includegraphics[width=0.99\linewidth]{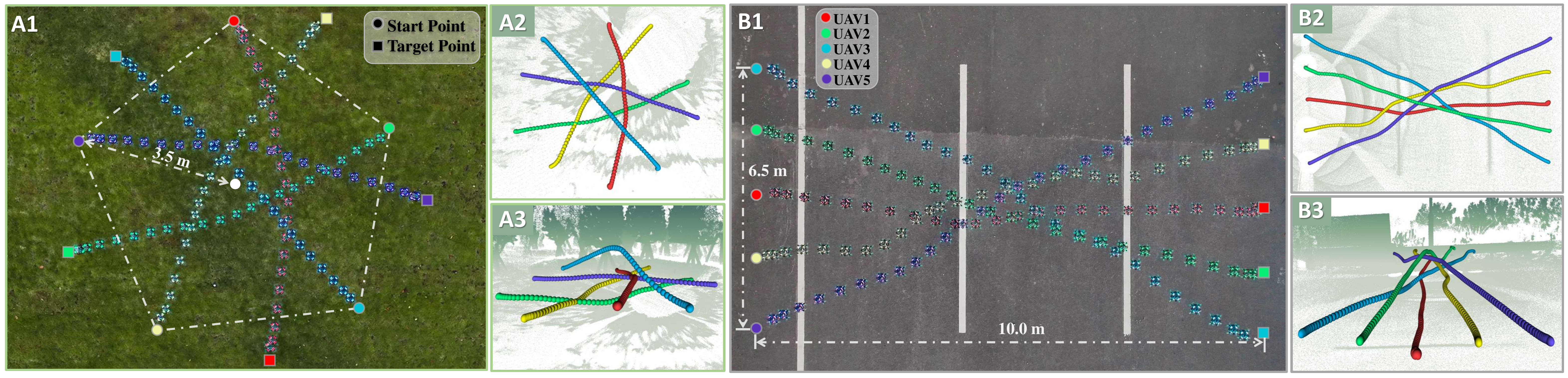}
	\caption{(A1,B1) The composite image of inter-UAV collision avoidance experiments, in which different colors represent different UAVs. Five UAVs first hover above the vertices of a regular pentagon (A1) or in a straight line (B1), and then accomplish collision-free flight using state estimation from Swarm-LIO2. (A2,A3,B2,B3)The estimated trajectories are visualized in different colors.} 
	\label{fig:mutual_avoidance} 
  \vspace{-3mm}
\end{figure*} 

This experiment emulates a dense air traffic scenario by flying five UAVs in interleaved directions (Fig.~\ref{fig:mutual_avoidance}(A1,B1)). Two flight tasks are demonstrated: in the first one, five UAVs initially hovering at the five vertices of a pentagon need to fly to a target position on the opposite side of the pentagon. In the second one, five UAVs initially hovering on one side of a field need to reach the other side of the field, meanwhile interchanging their positions. In both tasks, Swarm-LIO2 serves as the infrastructure for swarm initialization and swarm state estimation, which provides accurate global extrinsic transformations and real-time mutual state for inter-UAV collision avoidance. The inter-UAV collision avoidance is achieved by a swarm planner modified from \cite{zhou2022swarm,zhou2021ego}. The planned trajectories are fed into the motion controller\cite{lu2022model} for execution.

The composite snapshots illustrating the entire flight process are shown in Fig.~\ref{fig:mutual_avoidance}(A1,B1), with the estimated trajectories of each UAV shown in Fig.~\ref{fig:mutual_avoidance}(A2,A3,B2,B3). 
It can be observed that the estimated trajectories highly match the actual flights in the composite snapshots, which qualitatively validate the accuracy of Swarm-LIO2. We also analyze the state estimation result quantitatively, which is demonstrated in Section \ref{section:quantitative}.

\begin{figure*}[tbp]
	\setlength\abovecaptionskip{-0.1\baselineskip}
	\centering
	\includegraphics[width=0.99\linewidth]{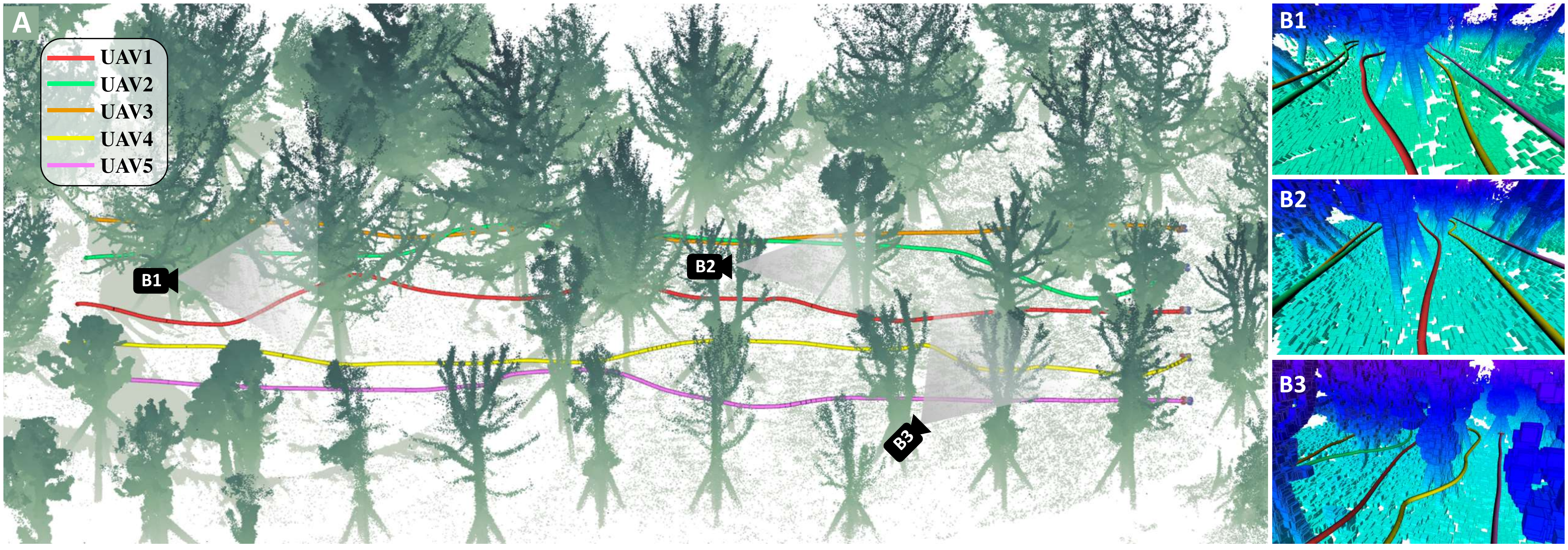}
	\caption{(A) a swarm system with 5 UAVs flying through a dense forest, the illustrated ego-state and the mutual state are estimated by UAV1. (B1-B3) Some details of the estimated trajectories when the UAVs avoid obstacles.}
	\label{fig:cover_rviz} 
  \vspace{-3mm}
\end{figure*}

\subsection{Fly Through a Dense Forest}\label{section:through_forest}
To validate the performance of Swarm-LIO2 in cases of mutual observation loss, we conduct a test in a dense forest environment using the five UAVs as shown in Fig.~\ref{fig:platform}. Each UAV needs to fly through a dense forest and reach each UAV's target point which is \SI{40}{m} away from the start point. During the whole process, no collision with obstacles in the environment or with teammate UAVs is allowed.

The initialization, goal transformation, and trajectory planning are the same as those in Section \ref{section:inter-UAV}.
Then all the UAVs start to fly through the forest from one side to the other, shown in Fig.~\ref{fig:cover_rviz}(A). 
During the flight, the dense trees lead to frequent mutual observation losses, while Swarm-LIO2 can still achieve robust and smooth mutual state estimation. The trajectories of the five UAVs and the point-cloud of the forest are depicted in Fig.~\ref{fig:cover_rviz}(A). The red trajectory represents the self-estimated flight trajectory of UAV1, while the green, orange, yellow, and purple trajectories represent the other UAVs' mutual state estimation results estimated by UAV1 in its respective global frame. Some details of the estimated trajectories when the UAVs avoid obstacles are illustrated in Fig.~\ref{fig:cover_rviz}(B1-B3).
Throughout the entire mission, Swarm-LIO2 provides accurate, real-time state estimation results (see the quantitative analysis in Section \ref{section:quantitative}) for the planning and control modules to achieve collision-free flights.

\subsection{Target Tracking with Dynamic Joining and Leaving}

\begin{figure*}[tbp]
	\setlength\abovecaptionskip{-0.03\baselineskip}
	\centering
	\includegraphics[width=0.99\linewidth]{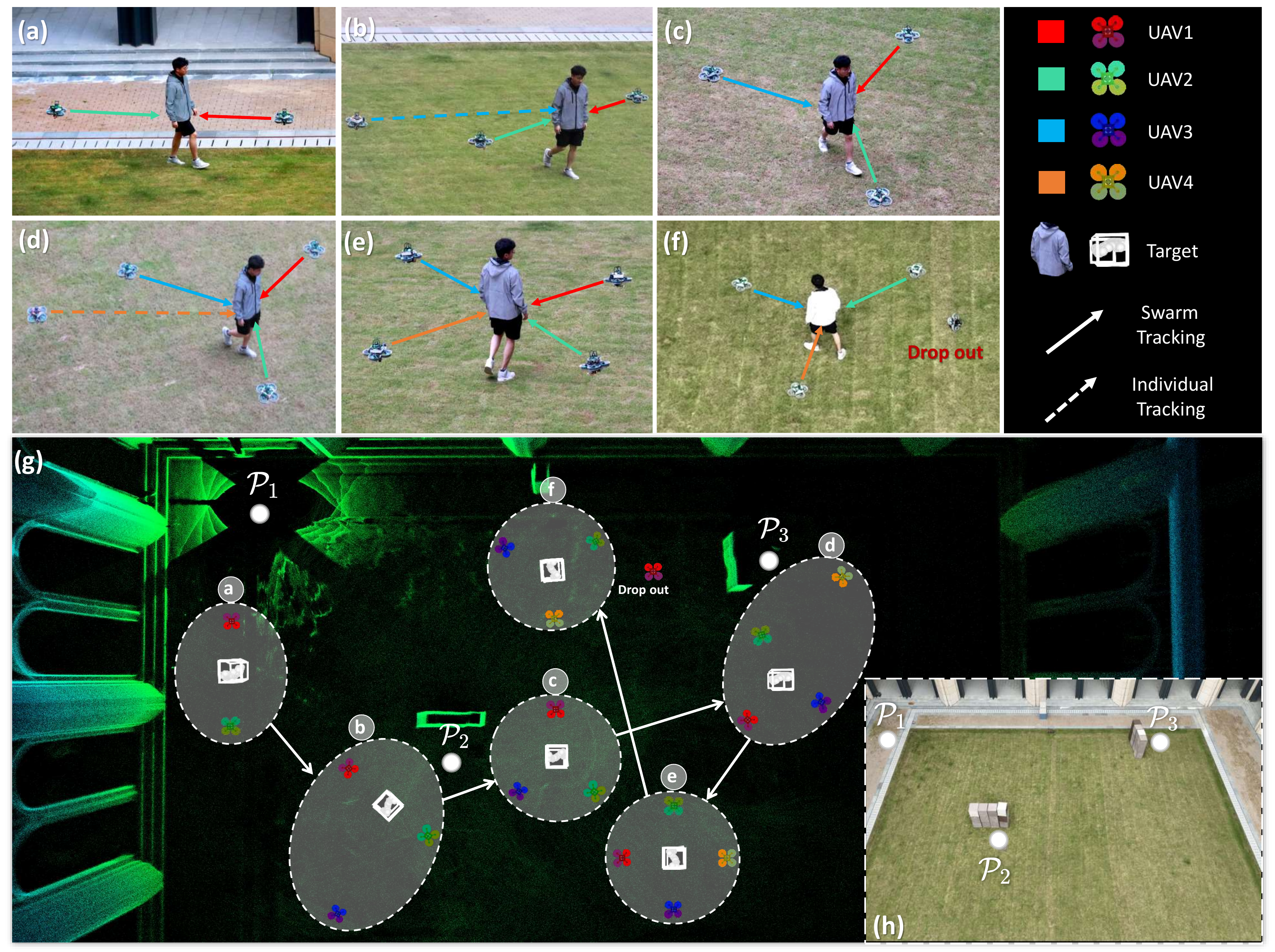}
	\caption{Collaborative target tracking experiment in an outdoor environment. Each UAV in the swarm system estimates its own and teammate UAVs' states by Swarm-LIO2. (a) The swarm only contains two members who track the target in a straight line. (b) UAV3 detects the target and tracks it in the individual tracking mode, and avoids other UAVs by treating them as dynamic obstacles. (c) UAV3 successfully joins the swarm after the online initialization, the formation changes into a triangle. (d) UAV4 detects the target, tracks the target, and tries to join the swarm. (e) UAV4 successfully joins the swarm, the formation changes to a square. (f) UAV1 is shut down intentionally. The formation transits back to a triangle, demonstrating the robustness of Swarm-LIO2 to single-point-of-failure. It is noted that different from the previous pictures captured by a ground camera, this picture is taken from the air, which has a different color for the person's coat due to different camera parameters. (g) The top-down view of the experiment site and the illustration of the entire swarm tracking application in Rviz. (h) Aerial view of the experiment site.} 
	\label{fig:SwarmTracking} 
  \vspace{-3mm}
\end{figure*}


To validate the plug-and-play property of Swarm-LIO2 which supports dynamic teammates joining and leaving, we conducted a collaborative target-tracking experiment with four UAVs. To enable fast detection of the target, the person being tracked wears a high-reflectivity vest, so his position can be easily detected from the high-reflectivity points from each UAV's LiDAR measurements. All UAVs have the same pre-programmed task: detecting and tracking a target, characterized by its high reflectivity and certain size, in a collaborative manner with teammates (if any) to maximize the overall target visibility, meanwhile avoiding the static and dynamic obstacles in the environment. In this process, Swarm-LIO2 serves as an infrastructure for automatic teammate finding, identification, and mutual state estimation, while trajectory planning is achieved by a decentralized swarm tracker in our previous work \cite{swarm_tracking}.  

Before the mission starts, UAV1 and UAV2 are placed at the same area $\mathcal P_1$ as shown in Fig.~\ref{fig:SwarmTracking}(g,h) where they can communicate well and are commanded to complete the swarm initialization by flying one UAV along a figure-8 trajectory. After the initialization, the two UAVs form a swarm system of size two. UAV3 and UAV4 are placed separately in different locations $\mathcal P_2$ and $\mathcal P_3$ respectively where they can't detect the target due to occlusion. 

The mission starts when the target enters the area $\mathcal P_1$, where UAV1 and UAV2 successfully detect the target and start to track it collaboratively. To maximize the target visibility, the two UAVs form a straight line with the target in the middle, as shown in Fig.~\ref{fig:SwarmTracking}(a). UAV3 and UAV4 are actively searching for the target but did not find one due to occlusions and FoV limit, hence they remain at their respective initial position. 

Subsequently, the target moves to the area $\mathcal P_2$, where it is detected by UAV3. Then UAV3 takes off and starts to track the target. Since UAV3 is not yet part of the swarm (it has neither been identified as a teammate by UAV1 and UAV2 nor the global extrinsic transformations are calibrated), it tracks the target in a solo manner by treating UAV1 and UAV2 as dynamic obstacles to avoid, as shown in Fig.~\ref{fig:SwarmTracking}(b). Similarly, UAV1 and UAV2 remain in their current formation, which is the optimal one for the tracking mission, while treating UAV3 as a dynamic obstacle to avoid. As a consequence, the swarm of UAV1 and UAV2 and the individual UAV3 both execute their pre-programmed task, although not in a collaborative manner due to the lack of a prior initialization.

As the tracking goes on, the UAV3 would fly a trajectory, during which its identity, temporal offset, and global extrinsic w.r.t. any of UAV1 and UAV2 can be estimated by Swarm-LIO2 on the fly. With the online initialization, UAV3 joins the swarm, forming a swarm system of size three. The new swarm system starts changing its form into a triangle shape, with the target in the center, to maximize the tracking visibility (see Fig.~\ref{fig:SwarmTracking}(c)). This process takes place automatically without interrupting the tracking task. When the target approaches area $\mathcal P_3$ where UAV4 is placed, the swarm size increases further to four and the UAVs form the shape of a square for maximizing the tracking visibility (see Fig.~\ref{fig:SwarmTracking}(d, e)). 

At last, UAV1 is killed intentionally to emulate a scenario in which one agent in the swarm experiences failures. Swarm-LIO2 on each remaining UAV can detect the teammate dropout, update its teammate list, and estimate the rest of teammate states all automatically, indicating that Swarm-LIO2 is robust to single-point-of-failure (Fig.~\ref{fig:SwarmTracking}(f)). Correspondingly, the planner quickly transforms the formation back into a triangle shape (Fig.~\ref{fig:SwarmTracking}(f)).

To sum up, in the entire target-tracking process, Swarm-LIO2 can conduct initialization on the fly, discover newly joined teammates or dropout teammates dynamically, and estimate the ego-state and mutual state in real-time, all take place automatically without interrupting the tracking task. This enables the swarm to adapt its formations to optimize task completion. Moreover, Swarm-LIO2 can provide consistent and accurate state estimation results throughout the entire mission (see quantitative results in Section \ref{section:quantitative}), assuring excellent target-tracking performance. To validate that Swarm-LIO2 possesses applicability to various environments, we also experimented with an indoor setting and a low-light night setting (see the attached video at \href{ https://youtu.be/Q7cJ9iRhlrY}{\textcolor{black}{https://youtu.be/Q7cJ9iRhlrY}}).

\color{black}
\subsection{More Experiments}
We conduct two more real-world experiments to highlight the superior robustness and broad applicability of Swarm-LIO2. In the first experiment, two UAVs equipped with different LiDARs fly in a degenerated scenario. With mutual observation constraints provided by Swarm-LIO2, the two UAVs can achieve centimeter-level localization accuracy. In the second experiment, the accurate mutual state estimation and global extrinsic calibration capability of Swarm-LIO2 enables three UAVs to transport a payload cooperatively from an outdoor scene into a building.
 Due to the page limit, we put the detailed descriptions, illustrations, and quantitative results in the Supplementary
Material \cite{zhu2024supplementary}.

\color{black}
\subsection{Time Consumption Analysis}
\begin{table}[tbp]
\centering
    \renewcommand\arraystretch{1.3}
    \caption{Average Time Consumption Per Scan (ms)}
    \label{table:time_real}
    \scalebox{0.85}{
    \begin{tabular}{p{1.7cm}<{\centering} p{0.7cm}<{\centering} p{0.7cm}<{\centering} p{0.7cm}<{\centering} p{0.8cm}<{\centering} p{0.9cm}<{\centering} p{0.9cm}<{\centering}}
    \toprule
        Method  & VII-B  &  VII-C &  VII-D &  \textcolor{black}{Sup-II}  &  \textcolor{black}{Sup-III} & Average\\
        \hline
        FAST-LIO2 & 5.28 & 5.49 & 7.23 &  8.10$^\dag$ & 5.31 & 6.28\\ 
        \textbf{Swarm-LIO2} & 6.74  & 6.87 & 8.33 &  9.13 & 6.76 & 7.57\\
        Swarm-LIO  & 10.29 & 10.83 &  11.36& 10.72 &  7.02 &  10.04\\
        LILI-OM  &  $\times$ & $\times$ &  26.36 &   32.68$^\dag$ & 22.35 & 27.13  \\
        \hline
    \end{tabular}
    }
    \begin{tablenotes}
        \footnotesize
        \item
        $^\dag$ denotes the case where UAV2 (equipped with s small FoV LiDAR) fails due to LiDAR degeneration, thus the time consumption is obtained from UAV1 only for single-LiDAR LIO methods.\\
        \item
        $\times$ denotes that LiLi-OM fails to extract enough features for optimization in the forest scene.        
     \end{tablenotes}
    \vspace{-2mm}
\end{table}

In this section, we evaluate the average computational time per LiDAR scan (unit: ms) of the aforementioned real-world experiments (from Section \ref{section:inter-UAV} to \textcolor{black}{Supplementary \cite{zhu2024supplementary}}), tested on the onboard computer NUC equipped with an Intel i7-1260P CPU. 
We compare the time consumption of Swarm-LIO2 with FAST-LIO2\cite{xu2022fast} (an efficient single-LiDAR LIO system), LiLi-OM\cite{liliom} (an optimization-based single-LiDAR LIO system) and our previous work Swarm-LIO \cite{zhu2023swarm}.
The average computational time of different methods in the different experiments is shown in Table~\ref{table:time_real}.

As can be seen, Swarm-LIO2 improves the computation efficiency significantly when compared to Swarm-LIO, due to the introduced state marginalization. 
In the experiment \textcolor{black}{shown in Supplementary III \cite{zhu2024supplementary}}, the time consumption of Swarm-LIO2 and Swarm-LIO are at a similar level because the three UAVs need to remain close to each other, leading to full mutual observation during the entire process. Therefore, the reduction in computation time caused by marginalization is not significant, and the slight reduction is primarily attributed to the more efficient point clustering algorithm FEC\cite{cao2022fec}.
However, in the other four experiments, since the mutual observation is frequently lost due to occlusions, the computational time of Swarm-LIO2 is obviously less than that of Swarm-LIO mainly due to the proposed marginalization operation. Compared to LiLi-OM, Swarm-LIO2 consumes much less time since it avoids time-consuming feature extraction and utilizes the efficient ESIKF framework. Compared to FAST-LIO2, despite Swarm-LIO2 incorporating many additional modules and handling more complex problems, it only incurs approximately 20$\%$ more computation time on average. 

Finally, since the LiDAR scan rate is \SI{30}{Hz}, indicating the limit of real-time computation is about \SI{33.33}{ms} per frame. In all the real-world experiments, the computational time of Swarm-LIO2 is far less than the limit value, showcasing the excellent real-time performance of Swarm-LIO2.

\subsection{Quantitative Analysis of State Estimation}\label{section:quantitative}

\begin{table}[tbp]
\centering
    \renewcommand\arraystretch{1.4}
    \caption{Standard Deviation of All Estimated States}
    \label{table:standard_deviation}
    \scalebox{0.85}{
    \begin{tabular}{p{0.5cm}<{\centering} p{1.7cm}<{\centering}   p{0.7cm}<{\centering}   p{0.7cm}<{\centering}  p{0.7cm}<{\centering}  p{0.8cm}<{\centering} p{0.9cm}<{\centering}  p{0.8cm}<{\centering}}
    \toprule
         Error & Method & VII-B & VII-C & VII-D & \textcolor{black}{Sup-II}  & \textcolor{black}{Sup-III}  & Average\\ 
          \cline{1-8}
        \multirow{2}* {\makecell[c]{\textbf{Pos} \\ \textbf{(m)}} } &
        \textbf{Swarm-LIO2}  &0.0539          &    \textbf{0.0515}  &   \textbf{0.0349}    &     \textbf{0.0256 } &0.0564&  \textbf{0.0445}  \\
                 & Swarm-LIO & \textbf{0.0537}&    0.0518           &    0.0383   &    0.0291   &    \textbf{0.0553}   &      0.0456              \\ 
        \multirow{2}* {\makecell[c]{\textbf{Rot} \\ \textbf{(rad)}} } &
        \textbf{Swarm-LIO2} &  \textbf{0.0631} & \textbf{0.0626}  & \textbf{0.0570} &  \textbf{0.0335}&   0.0698         & \textbf{0.0572}\\
                & Swarm-LIO &  0.0632          &    0.0629        &        0.0596   &       0.0352     & \textbf{0.0670}  & 0.0576\\
        \bottomrule
    \end{tabular}
    }
     \vspace{-2mm}
\end{table}

In this section, we quantitatively evaluate the state estimation consistency of Swarm-LIO2 in the aforementioned real-world
experiments. Since in most real-world experiments, no ground-truth of UAVs' states can be obtained, for a swarm system containing $N$ UAVs, we compute the standard deviation of all the $N^2$ estimated UAV trajectories in each experiment to quantitatively evaluate the state estimation consistency. 

The computed standard deviations of rotation and translation estimated by Swarm-LIO2 and Swarm-LIO\cite{zhu2023swarm} are illustrated in Table~\ref{table:standard_deviation}. As can be seen, the standard deviation of position and rotation is at the centimeter level and the degree level, respectively, indicating the excellent consistency of the swarm state estimation (both ego and mutual) of Swarm-LIO2. From the comparison with Swarm-LIO, it is evident that the two methods have similar state estimation consistency, indicating that the marginalization operations adopted in Swarm-LIO2 nearly do not impact the performance.

\color{black}
\subsection{Communication Bandwidth Analysis}
We quantitatively evaluate the data transfer volume (TX and RX) per second of each UAV in the aforementioned real-world experiments. The results are shown in Table~\ref{table:bandwidth}.  Both systems have extremely low average bandwidth usage, which is under \SI{35}{KB/s} when the swarm system contains 5 UAVs. 
In all the real-world experiments of Swarm-LIO2, the adopted wireless network adapter on each UAV is Intel Wi-Fi 6E AX211 (Gig+) with \SI{2.4}{Gbps} (approximately equals to \SI{300}{MB/s}) bandwidth, four orders of magnitude higher than the actual usage. Compared with Swarm-LIO, Swarm-LIO2 consumes a slightly larger bandwidth since it exchanges more information, including global extrinsic transformations and degeneration status.

We also calculated the packet loss rate (PLR) of the two methods. The average PLR is 11.08\% for Swarm-LIO2 and 10.36\% for Swarm-LIO, which are similar. The accurate state estimation as shown previously in the presence of the packet loss indicates the excellent robustness of our system.

\begin{table}[tbp]\color{black}
\centering
    \renewcommand\arraystretch{1.3}
    \caption{Average Communication Bandwidth Usage}
    \label{table:bandwidth}
    \scalebox{0.85}{
    \begin{tabular}{p{0.8cm}<{\centering} p{1.6cm}<{\centering} p{0.8cm}<{\centering} p{0.8cm}<{\centering} p{0.8cm}<{\centering} p{0.8cm}<{\centering} p{0.9cm}<{\centering}}
    \toprule
      \multirow{2}* {\makecell[c]{Method}}& Exp  & VII-B  &  VII-C &  VII-D &  Sup-II  &  Sup-III \\
        & Swarm Size & 5 & 5 & 4 & 2 &3\\
       \hline
        \multirow{2}* {\makecell[c]{\textbf{Swarm-} \\ \textbf{LIO2}}}& TX(KB/s) & 31.98  & 33.89 & 22.85 &  8.67 & 16.33 \\
                                                                    &RX(KB/s) &  27.97 &  27.43 & 19.75  &  7.94  & 14.97 \\
      \multirow{2}* {\makecell[c]{Swarm- \\ LIO}} & TX(KB/s)  & 25.54 & 25.98 &  18.48 & 6.52 &  14.52\\
                                                &RX(KB/s)  &  21.86 & 21.32  &  15.66  & 5.93  &  12.89 \\
        \hline
    \end{tabular}
    }
    \vspace{-2mm}
\end{table}

\color{black}
\section{Conclusion and Future Work}
In this paper, we proposed \textbf{Swarm-LIO2}, a decentralized, efficient state estimation framework based on LiDAR and IMU measurements for aerial swarm systems. A decentralized temporal calibration approach was utilized to calibrate the inter-UAV temporal offset.
Novel reflective tape-based UAV detection, trajectory matching, and factor graph optimization-based methods were proposed to perform efficient and fast teammate identification and global extrinsic calibration. A novel marginalization module was proposed to reduce the state dimension and further improve the swarm scalability, and a degeneration evaluation module was presented to ensure robust ego-state determination.
Furthermore, we introduced the elaborate measurement modeling and temporal compensation of the mutual observation measurements, enhancing our state estimator's accuracy and consistency. By exchanging bandwidth-efficient information via an Ad-Hoc network, the mutual observation measurements are tightly coupled with IMU and point-cloud measurements under an ESIKF framework, fulfilling real-time, accurate ego-state mutual state estimation. Using simulation benchmarks, we compared our LiDAR-inertial odometry with other state-of-the-art LIO methods, demonstrating excellent robustness to LiDAR degenerated scenarios. \textcolor{black}{In addition, the analysis of computational time and communication bandwidth usages at different swarm scales showcases the superior scalability of our method.}
Besides, we integrated our method into a UAV swarm composed of at most five UAVs with fully autonomous state estimation, planning, and control modules.
Various simulated and real-world experiments were conducted, demonstrating that our method serves as an infrastructure for aerial swarm systems and can support a wide range of UAV swarm applications.

In the future, we will focus on extending Swarm-LIO2 to a more complete swarm SLAM system by incorporating loop closure modules and historical pose correction, to ensure low drift after a long time of running.

\section*{Acknowledgement}
The authors thank Ms. Yang Jiao, Mr. Wendi Dong, and Mr. Meng Li for the helpful discussion, and Prof. Ximin Lyu for the experiment site support. The authors acknowledge funding from CETC and DJI, and equipment support from Livox Technology.

\bibliographystyle{support/IEEEtran}
\bibliography{paper}

\clearpage
\twocolumn[
\begin{@twocolumnfalse}
\section*{
    \begin{center}
        \textbf{\Large Supplementary Materials for Swarm-LIO2: Decentralized, Efficient LiDAR-inertial Odometry for UAV Swarms\\[25pt]}
    \end{center}
 }
\end{@twocolumnfalse}
]

\section{Fly Through a Degenerated Corridor}

\begin{figure}[htbp]
\setlength\abovecaptionskip{-0.05\baselineskip}
	\centering
	\includegraphics[width=0.99\linewidth]{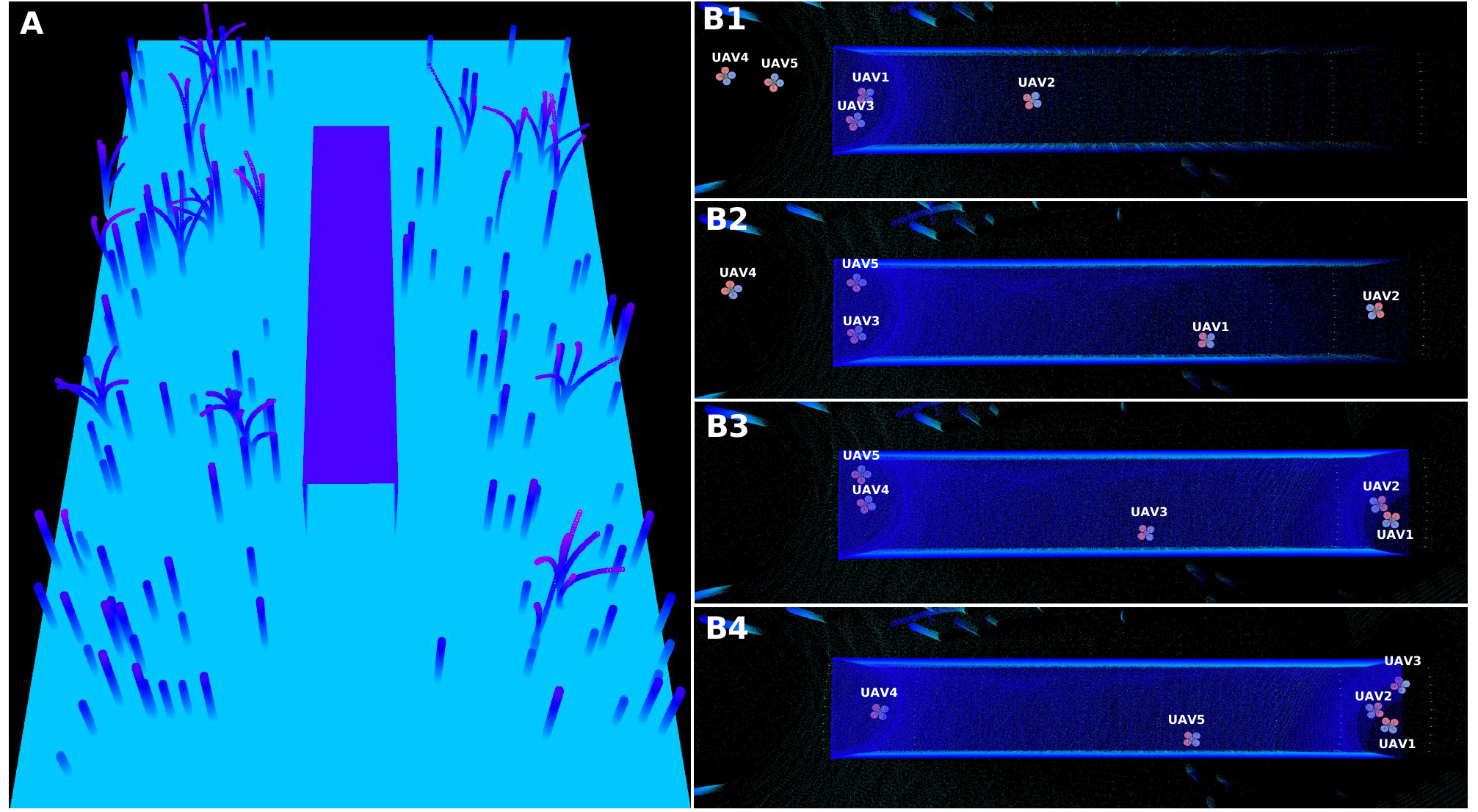}
	\caption{(A) The simulation environment of the degenerated corridor. (B1-B4) The UAVs fly through the corridor sequentially, so that the mutual observation measurements can be leveraged for localization of UAVs in the corridor.} 
	\label{fig:through_corridor} 
\end{figure}

In this section, we conduct a simulation experiment in which five UAVs equipped with Livox Mid360 LiDARs need to fly through a degenerated corridor (Fig.~\ref{fig:through_corridor}). In this case, the measurements of a single LiDAR can not provide sufficient constraints for pose determination, but Swarm-LIO2 can perform robust and stable state estimation thanks to the mutual observation measurements from teammates. 

\begin{figure}[htbp]
	\setlength\abovecaptionskip{-0.05\baselineskip}
	\centering
	\includegraphics[width=0.99\linewidth]{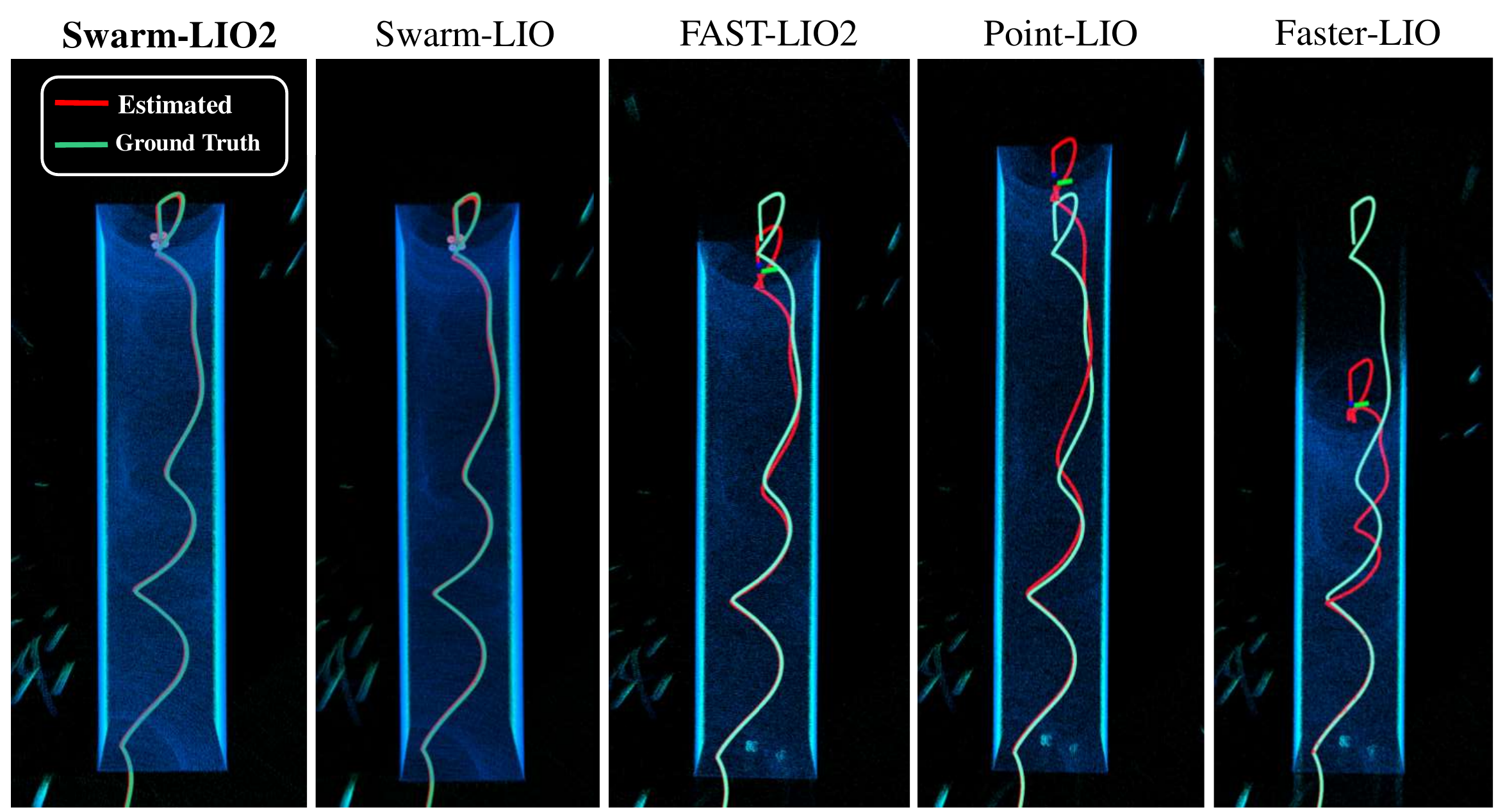}
	\caption{Bird eye view of the constructed point-cloud map and the trajectory of UAV2 estimated by Swarm-LIO2, Swarm-LIO, FAST-LIO2, Point-LIO, and Faster-LIO.} 
	\label{fig:degrade_comparison} 
\end{figure}

In the simulation, the UAVs fly cooperatively through the corridor one by one. When the first UAV, here is UAV2, flies into the corridor, other UAVs hover at the entrance (where sufficient structural features exist for their pose estimation) and provide mutual observation measurements for UAV2. As shown in Fig.~\ref{fig:through_corridor}(B1), when UAV2 detects LiDAR degeneration, it leverages mutual observation measurements from the rest UAVs to achieve robust state estimation. Then UAV2 flies through the corridor and hovers at the end of the corridor, offering mutual observation measurements for the rest UAVs to pass through the corridor (Fig.~\ref{fig:through_corridor}(B2-B4)).

\begin{figure}[tbp]
	\setlength\abovecaptionskip{-0.05\baselineskip}
	\centering
	\includegraphics[width=0.99\linewidth]{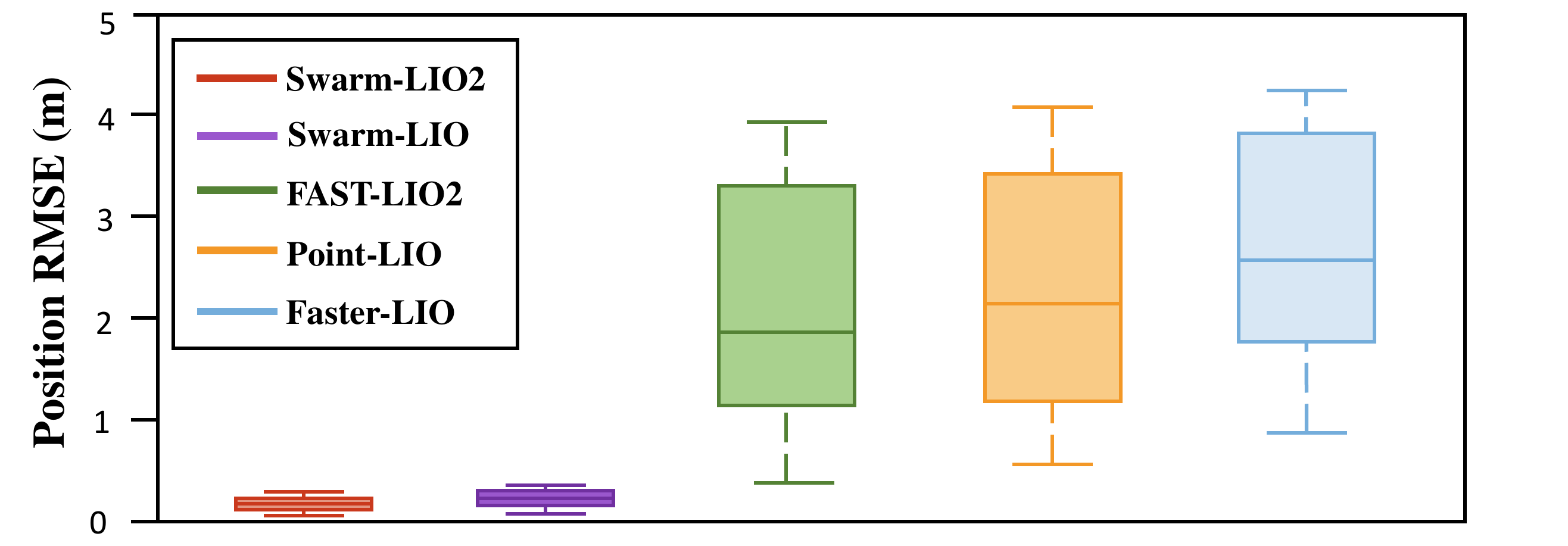}
	\caption{The localization accuracy comparison in a degenerated corridor scene among Swarm-LIO2, Swarm-LIO, FAST-LIO2, Point-LIO, and Faster-LIO.} 
	\label{fig:boxplot}
\end{figure}

 As far as we know, apart from our previous work Swarm-LIO \citesec{zhu2023swarm}, there is no other open-sourced 3D LiDAR-based state estimation method for UAV swarm, thus we compare the localization accuracy of our method to Swarm-LIO and some state-of-the-art LiDAR-inertial odometry for a single UAV system, including FAST-LIO2\citesec{xu2022fast}, Point-LIO\citesec{he2023point}, and Faster-LIO\citesec{bai2022faster}, in which the ground-truth of the ego-state is provided by the simulator. 
Take UAV2 as an example, the point-cloud map, the self-estimated position trajectory, and the ground-true trajectory are illustrated in Fig.~\ref{fig:degrade_comparison} with quantitative results supplied in Fig.~\ref{fig:boxplot}. It can be observed that in degenerated scenes, by fusing mutual observation measurements, the localization errors of Swarm-LIO2 and Swarm-LIO are much smaller than those of other single-agent LiDAR-inertial odometry methods. Moreover, in such a degenerated scenario, Swarm-LIO2 can achieve slightly better self-localization robustness and accuracy than Swarm-LIO, which is mainly attributed to the careful measurement modeling (Section V-C1) and temporal compensation (Section V-C2) in the paper.

\section{Flight In Degenerated Scenario}\label{section:degenerated}

\begin{figure*}[tbp]
	\setlength\abovecaptionskip{-0.05\baselineskip}
	\centering
	\includegraphics[width=\linewidth]{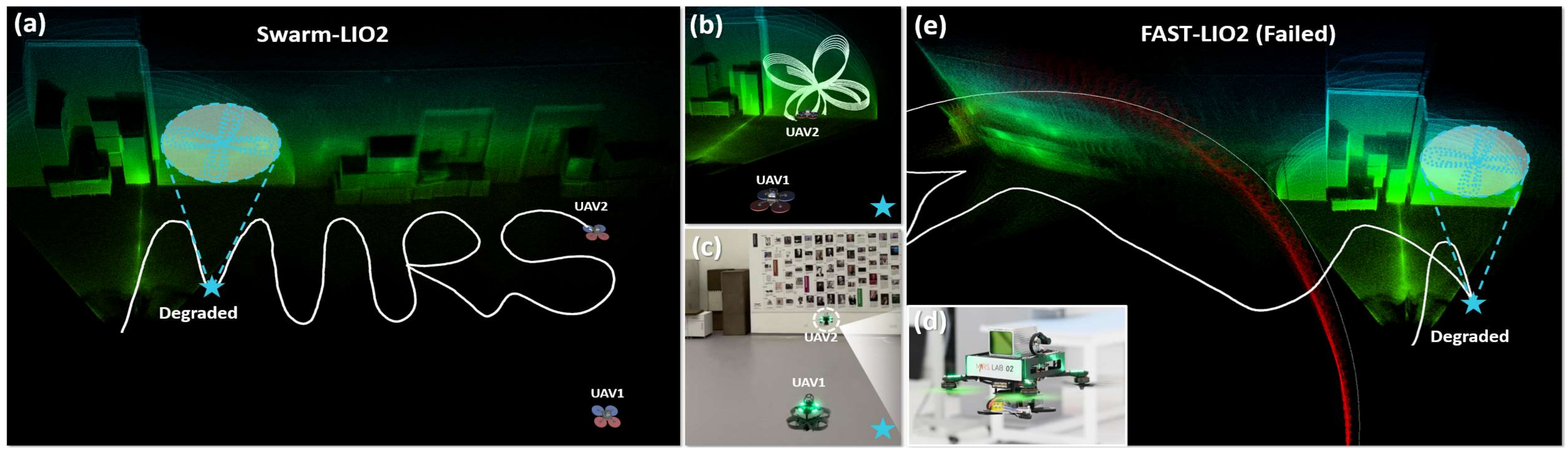}
\caption{Degenerated experiment of facing a smooth plane. (a) Clear and consistent point-cloud map constructed and state estimated by Swarm-LIO2 fusing passive observation measurements ${^{b_1}}\mathbf p_{b_2}$. The blue star represents the degenerated pose of UAV2 where the LiDAR is facing a smooth wall. UAV1 is flying behind UAV2 to provide mutual observation measurements. (b) Detail of the degenerated pose, the white points represent the current LiDAR scan. (c) Third personal view of the degenerated pose. (d) Detailed depiction of UAV2, a quadrotor equipped with Livox Avia LiDAR. (e) Point-cloud map constructed and state estimated by FAST-LIO2, severe drift occurs due to scarce structural constraints, resulting in a messy map.} 
	\label{fig:degrade} 
\end{figure*}

To validate the robustness of Swarm-LIO2 in LiDAR degenerated environments, we conduct a flight experiment for a swarm consisting of two UAVs. UAV1 carries a Livox Mid360 LiDAR, while UAV2 carries a Livox Avia LiDAR with a smaller FoV (especially in the horizontal direction) which is only 70.4$^\circ$ $\times$ 77.2$^\circ$, shown in Fig.~\ref{fig:degrade}(d). We instructed UAV2 to follow a pre-planned trajectory, shaped like ``\textit{MARS}" which is the name of our laboratory. At certain poses, the Avia LiDAR mounted on UAV2 would directly face a smooth plane, leading to LiDAR degeneration. During the entire flight, UAV1 is flying behind UAV2 as an observer to provide UAV2 passive mutual observation measurements.

\begin{figure}[tbp]
\setlength\abovecaptionskip{-0.05\baselineskip}
	\centering
	\includegraphics[width=0.99\linewidth]{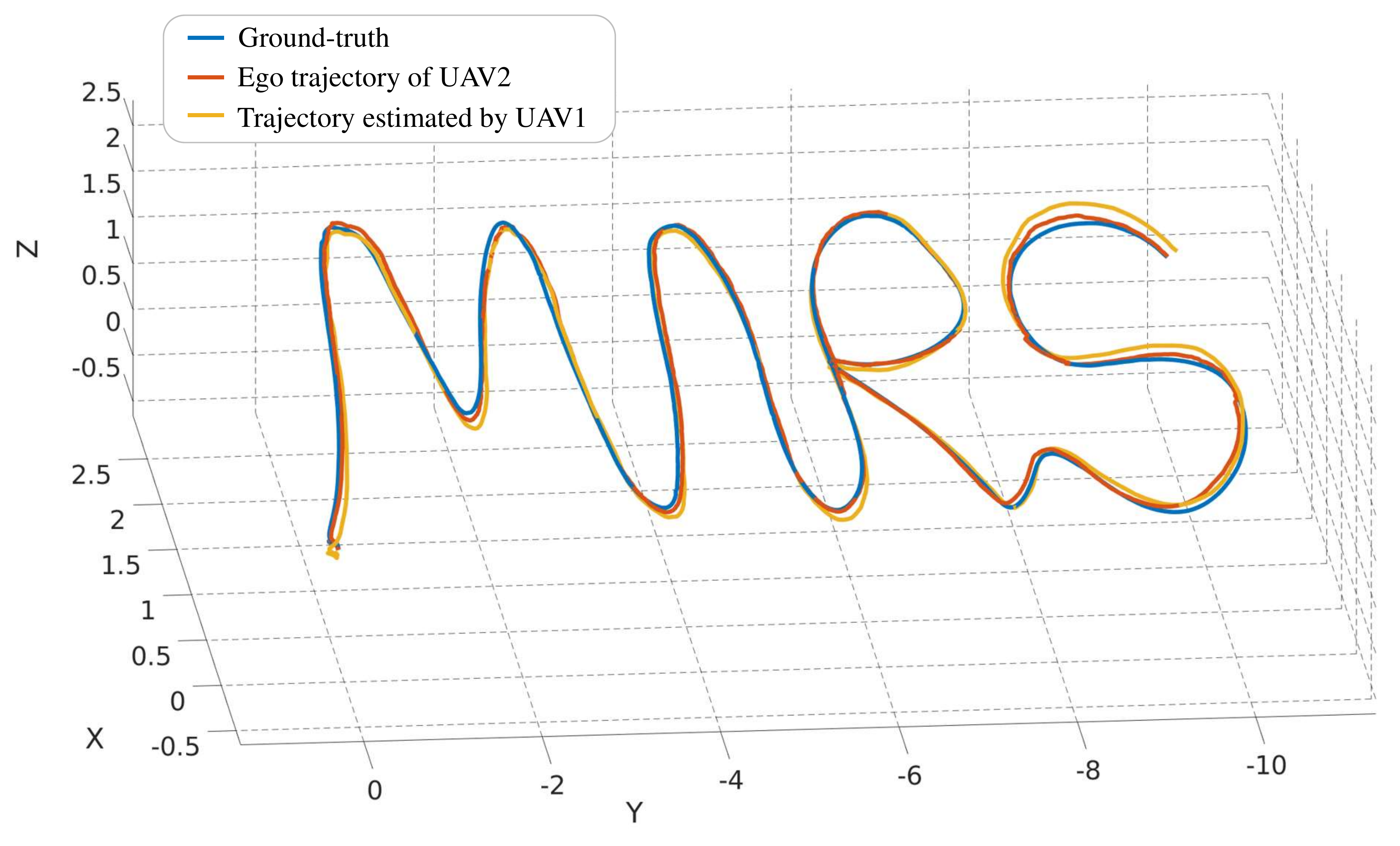}
	\caption{Trajectory of UAV2 estimated by itself, UAV1, and the motion capture system (\ie, the ground-truth).} 
	\label{fig:mars_traj} 
\end{figure}

We compare the self-localization result of our method with a representative single-agent LIO system, FAST-LIO2\citesec{xu2022fast}, as shown in Fig.~\ref{fig:degrade}. Since the LiDAR measurements can not provide sufficient constraints for pose determination, the state estimation of FAST-LIO2 diverges soon, the odometry largely drifts, and the point-cloud map gets messy. For Swarm-LIO2, the passive observation measurements offered by UAV1 provide the necessary information for robust localization and a consistent map. For the degenerated UAV2. The ground-truth provided by the motion capture system, the self-estimated trajectory of UAV2, and the trajectory of UAV2 estimated by UAV1 are depicted in Fig.~\ref{fig:mars_traj}. The average position error estimated by UAV2 itself is \textbf{0.043m} and that estimated by UAV1 is \textbf{0.059m}, both in the centimeter level.

\section{Cooperative Payload Transportation}\label{section:payload}

\begin{figure}[tbp]
\setlength\abovecaptionskip{-0.05\baselineskip}
	\centering
	\includegraphics[width=0.99\linewidth]{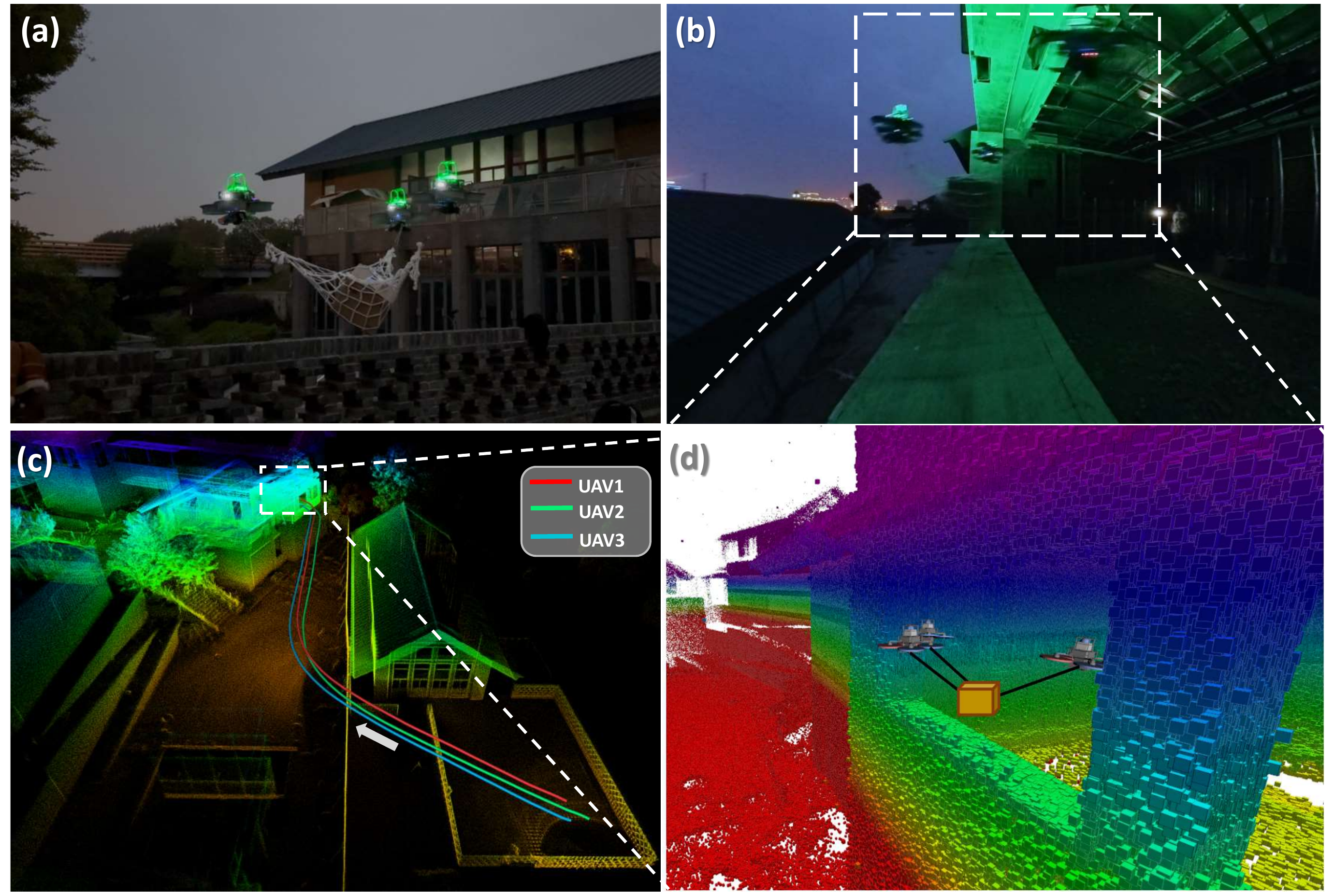}
	\caption{Cooperative Load Transportation in Outdoor Scenario. (a) Picture of the payload suspended swarm flying in the outdoor scenes. (b) Picture of the swarm flying through the window.(c) The constructed point-cloud map, the planned path of UAV1, the take-off, and the target area. (d) The poses of the three UAVs and the point-cloud of the environment at the moment when fly through the window. } 
	\label{fig:Load} 
\end{figure}

In this section, we implemented an interesting application in which a swarm composed of three UAVs completes the payload transportation from an outdoor location to an indoor area, see Fig.~\ref{fig:Load}. The focus of this application is to demonstrate the capability of estimating the temporal offset and the global extrinsic transformations of Swarm-LIO2, rather than trajectory planning for the swarm. Therefore, in this experiment, the trajectory for UAV1 is priorly planned to ensure collision-free flight.
The trajectories of the other two UAVs are obtained by transforming online the pre-planned trajectory into their respective global frames with appropriate offsets, using the precise temporal offset and global extrinsic transformations provided online by Swarm-LIO2. The three trajectories are then tracked independently with controller \citesec{lu2022model}. 

After loading the payload, the three UAVs fly in a low-light, outdoor scenario shown in Fig.~\ref{fig:Load}(a), and ultimately enter a building through a window, shown in Fig.~\ref{fig:Load}(b). Throughout the entire flight, the three UAVs maintain the formation of a triangle, ensuring that each UAV contributes nearly equal pulling forces. It is accurate state estimation, temporal offset, and global extrinsic calibration that empowers the UAVs to maintain the correct formation at any given moment, and successfully complete the payload transportation mission without collisions. The point-cloud map of the whole experiment site, which is constructed in real-time, and the transformed paths of the three UAVs are illustrated in Fig.\ref{fig:Load}(c). The poses of the three UAVs and the point-cloud of the environment at the moment when UAVs were flying through the window are illustrated in Fig.\ref{fig:Load}(d). It is worth noting that the entire swarm successfully flies through the narrow window without any collisions between the UAVs, payload, and the surrounding environment.

\bibliographystylesec{support/IEEEtran}
\bibliographysec{supplementary}

\end{document}